\documentclass{article}

\PassOptionsToPackage{numbers,sort&compress}{natbib}
 \usepackage[preprint]{neurips_2026}

\usepackage[utf8]{inputenc} 
\usepackage[T1]{fontenc}    
\usepackage{hyperref}       
\usepackage{url}            
\usepackage{booktabs}       
\usepackage{amsfonts}       
\usepackage{nicefrac}       
\usepackage{microtype}      
\usepackage{xcolor}         
\usepackage{amsmath}
\usepackage{cleveref}
\usepackage{graphicx}
\usepackage{multirow}
\usepackage{booktabs}
\usepackage{subcaption}
\usepackage{wrapfig}
\usepackage{rotating} 

\usepackage{cleveref}
\crefname{section}{Sec.}{Secs.}
\Crefname{section}{Section}{Sections}
\crefname{equation}{Eq.}{Eqs.}
\Crefname{equation}{Equation}{Equations}
\crefname{figure}{Fig.}{Figs.}
\Crefname{figure}{Figure}{Figures}
\crefname{table}{Tab.}{Tabs.}
\Crefname{table}{Table}{Tables}
\crefname{section}{App.}{Apps.}
\title{Out-of-Distribution Detection in Continual Learning}

\author{%
  Nimeshika Udayangani$^{1}$ \quad
  Flora D.~Salim$^{2}$ \quad
  Christopher Leckie$^{1}$ \quad
  Sarah Erfani$^{1}$ \\[3pt]
  $^{1}$School of Computing and Information Systems,
  The University of Melbourne, Australia \\
  $^{2}$School of Computer Science Engineering, University of New South Wales, Australia \\
  \texttt{nhewadehigah@student.unimelb.edu.au}
}

\begin{document}

\maketitle

\vspace{-10pt}
\begin{abstract}
\vspace{-8pt}
Out-of-distribution detection in continual learning (OOD-CL) is highly challenging, as it requires (i) learning a sequence of tasks without catastrophic forgetting, and (ii) identifying unknown samples (OOD) without access to known (in-distribution (ID)) data from past tasks, where standard confidence-based OOD scores tend to degrade.
Existing methods largely rely on a rehearsal buffer of past data or assume closed-world settings, limiting their scalability and reliability in real-world scenarios. 
In this paper, we propose a novel OOD detection framework for rehearsal-free continual learning, inspired by Kolmogorov–Arnold Networks (KAN), which mitigate catastrophic forgetting via their inherent local neuroplasticity.
To better leverage KAN in OOD-CL, we introduce a task-adaptive classifier architecture, \emph{TA-KAN}, which mitigates recency bias and enhances ID--OOD separability through task-specific control of activation locality. 
Furthermore, we propose a geometry-guided OOD score that complements TA-KAN classifier confidence without relying on past ID data. 
Our method significantly improves OOD detection while maintaining strong ID classification performance, achieving state-of-the-art results in OOD-CL.

\end{abstract}


\section{Introduction}

Continual learning~(CL)~\cite{CL_survey, CL_survey2} aims to enable AI models to mimic human-level cognitive abilities by allowing them to learn continuously from evolving data streams. 
Despite extensive research in this area~\cite{CIL1st,CPrompt,Bisecle,L2P,ER,GEM}, most existing methods operate under the closed-world assumption, where test data are drawn from the observed training distribution (in-distribution, ID).
However, in real-world deployments, models inevitably encounter unknown inputs (out-of-distribution, OOD), often leading to overconfident and potentially misleading predictions~\cite{hendrycks_2017MSP,proKT}.
This limitation underscores the importance of robust OOD detection in continual learning systems (\textit{OOD-CL}) for trustworthy deployment in open-world scenarios.
In particular, OOD-CL is highly challenging, as it requires (i) learning a sequence of tasks without catastrophic forgetting—i.e., a significant deterioration in performance on previously learned tasks—and (ii) identifying OOD samples without access to known (ID) data of past tasks.
This necessitates the model to maintain reliable decision boundaries between inter-task classes as new classes are continuously introduced, while also preserving a clear separation between OOD samples and previously learned classes, which becomes particularly challenging for older ID classes.

To address the first challenge, widely adopted rehearsal-based CL~\cite{ER,GEM, syntheticOOD} stores a subset of past training data in a memory buffer to mitigate catastrophic forgetting.
However, such approaches have limited practical applicability due to privacy and memory constraints~\cite{L2P,Dual_prompt}.
In contrast, recent advancements in pre-trained models and large language models (LLMs) \cite{LLMs} have enabled prompt-based methods, which have emerged as state-of-the-art ~\cite{CODA_prompt,CPrompt,Bisecle,L2P,Dual_prompt} and provide a rehearsal-free alternative for CL.
Instead of directly modifying encoder weights on newly arriving data, these methods introduce a small set of learnable parameters (prompts) to instruct the model on sequential tasks, thereby reducing forgetting and encoding task-specific knowledge in a more succinct way.
However, they are largely developed and evaluated in closed-world settings.
Meanwhile, addressing OOD detection is non-trivial, as naively adopting existing methods that assume a static in-distribution is challenging and insufficient in the CL setting.
Standard OOD detection methods, which are largely based on the confidence scoring of the classifier (e.g., MSP~\cite{hendrycks_2017MSP}, entropy~\cite{shannonEntropy}, energy~\cite{EnergyOE}, MaxLogit~\cite{MaxLogit}), can degrade in the CL setting, as they may assign low confidence to ID samples due to the objective of improving generalizability for new data~\cite{OODCILunsuper}.
Moreover, the absence of past data makes a large class of OOD detection~\cite{lee_2018_mahalanobis, GramMatrix, retrainOOD, retrainOOD2, KNN_OOD, hendrycks_2018OE, syntheticOOD} non-scalable, as they rely on access to full training in-distribution or require modifications to the training procedure.
Nonetheless, the few works that attempt to address OOD-CL either naively couple standard OOD methods with CL models~\cite{MORE1, proKT, ORE} or still assume access to past data~\cite{MORE, H2ST, ORE}.
In contrast, in this paper, we propose a novel method for OOD detection in rehearsal-free continual learning.

\begin{wrapfigure}{r}{0.4\linewidth}
\vspace{-10pt}
    \centering
    \includegraphics[width=\linewidth]{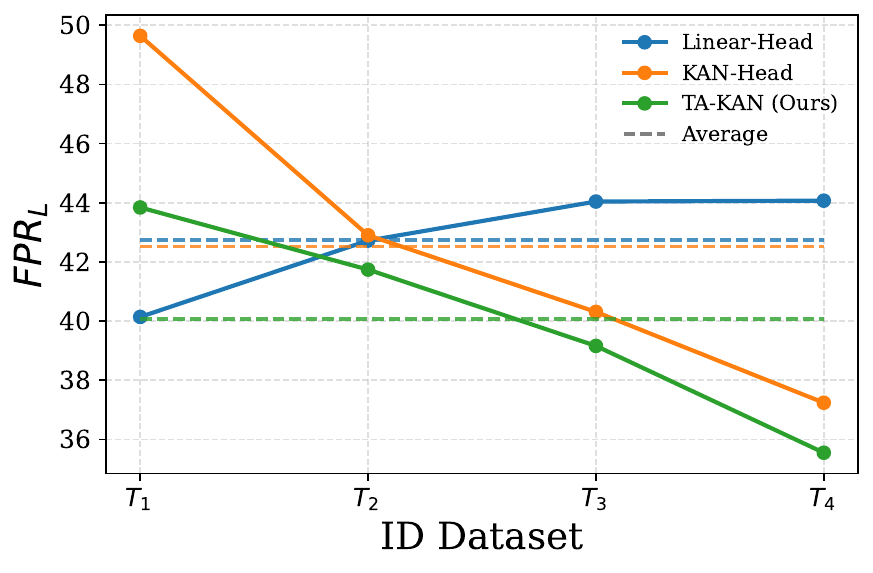}
    \caption{FPR of a trained CIL model on ImageNet-R (5-task), evaluated across task-wise ID settings. KAN-Head exhibits significantly higher FPR when older tasks are treated as ID compared to recently learned tasks, revealing a bias toward new-task classes. The proposed TA-KAN mitigates this bias, thereby reducing the average FPR.}
    \label{fig:bias_plot}
    \vspace{-10pt}
\end{wrapfigure}

\looseness=-1
Our study is motivated by the recent success of Kolmogorov–Arnold Networks (KAN)~\cite{KAN} in mitigating catastrophic forgetting, owing to their distinctive architecture. 
Unlike traditional multi-layer perceptrons (MLP), KAN incorporates trainable activation functions instead of shared weights with fixed activations, thereby exhibiting the property of local neuroplasticity~\cite{KAN} (see \cref{fig:kan_vs_mlp} in the Appendix for an illustration).
This suggests that learning a new task primarily affects only the regions of the network relevant to the current input, while leaving distant regions largely unaffected.
Hence, KANs are naturally effective in mitigating catastrophic forgetting~\cite{KAN} and consistently outperform their linear counterparts in the closed-world~\cite{KAC}. 
However, upon closer examination of KAN’s behavior in the open-world setting, we observe that OOD detection is heavily biased toward new classes (\cref{fig:bias_plot}).
In particular, KAN struggles to distinguish OOD samples from older ID classes, causing a large fraction of them to be misclassified as OOD and resulting in a high false positive rate.
This suggests that the increased ID performance comes at the expense of a bias toward new tasks.

To address this, we introduce a novel task-adaptive architecture (\emph{TA-KAN}) for RBF-KAN-based classifiers~\cite{KAC}, tailored for OOD-CL.
RBF-KANs~\cite{RBFKAN} replace the B-spline basis functions in original KANs~\cite{KAN} with radial basis functions (RBFs)~\cite{RBF} to better address the curse of dimensionality, and improve locality~\cite{RBFKAN2,KAC}.
Specifically, TA-KAN introduces a task-adaptive width selection for RBF bases, where each task is associated with a distinct width that defines its activation range of interest. 
Locality is then adaptively controlled based on task recency, improving separability between ID and OOD samples in more ambiguous regions, thereby improving overall OOD-CL performance.
Importantly, the proposed adaptive RBF classifier retains the same parameterization, preserving the parsimony required for CL~\cite{parsimony_CIL}.

Furthermore, to address the degradation of confidence-based scoring in the proposed classifier, we introduce complementary geometric guidance for its classifier derived OOD scores, thereby proposing a novel OOD score tailored for CL.
Since the KAN-based CL model is optimized for improving generalizability to mitigate catastrophic forgetting, this can also induce lower confidence for ID data compared to the static setting \cite{OODCILunsuper}, leading to increased overlap between ID and OOD scores, reducing their separability.
Specifically, traditional geometry-based OOD methods (e.g., KNN~\cite{KNN_OOD}, NNGuide~\cite{NNGuide}, LID~\cite{SupLID}) require access to the full training distribution. 
In contrast, our proposed guidance is rehearsal-free and operates in the prompt space, which encodes higher-level task semantics compared to the lower-level embedding space used by conventional methods. 
Therefore, it jointly considers both class-level and task-level manifolds, thereby introducing a two-level manifold constraint for OOD-CL. 
The resulting OOD score further enhances the performance of the TA-KAN classifier, achieving SOTA results by a large margin.

\noindent\textbf{We summarize our key contributions as follows.}
We propose (i) a novel OOD detection framework for rehearsal-free CL, addressing practical constraints such as memory, privacy, and reliability; 
(ii) a task-adaptive KAN classifier that better leverages the neuroplasticity of KAN in the open-world by mitigating recency bias and enhancing ID–OOD separability, while preserving parsimony; and 
(iii) a geometry-guided OOD scoring function that complements classifier confidence, thereby formulating OOD-CL as a two-level manifold constraint, without reliance on full in-distribution.

\section{Preliminaries and Notations} \label{sec:preliminary}


\textbf{Out-of-distribution (OOD) detection.}
Let $\mathcal{D}_{\text{in}} = \{(\boldsymbol{x}^i, y^i)\}_{i=1}^n$ denote the training dataset, 
where each input $\boldsymbol{x}^i \in \mathcal{X}$ is associated with a label 
$y^i \in \mathcal{Y} = \{1, 2, \dots, C\}$, with $C$ denoting the total number of classes. 
Let $f : \mathcal{X} \rightarrow \mathbb{R}^C$ be a neural network trained on $\mathcal{D}_{\text{in}}$.
At test time, the objective is twofold: (i) to correctly classify samples from known classes in $\mathcal{Y}$ (in-distribution (ID) classes), and (ii) to effectively identify samples that do not belong to any class in $\mathcal{Y}$, i.e., out-of-distribution (OOD) samples. 
To this end, a scoring function $S : \mathcal{X} \rightarrow \mathbb{R}$ is defined, where higher scores indicate a higher likelihood of a sample being in-distribution. Given a threshold $\eta \in \mathbb{R}$, the decision rule is defined as:
\begin{equation} \label{eq:ood_score}
\text{Decision}(\boldsymbol{x}) =
\begin{cases}
\text{ID}, & \text{if } S(\boldsymbol{x}) \geq \eta, \\
\text{OOD}, & \text{otherwise}.
\end{cases}
\end{equation}
The scoring function $S$ can be derived either from the classifier outputs $h(\boldsymbol{x})$ (e.g., confidence-based scores~\cite{hendrycks_2017MSP, EnergyOE, shannonEntropy, MaxLogit}) or from the geometry of the embedding space $g(\boldsymbol{x})$ \cite{KNN_OOD, NNGuide, SupLID, VIM}, where $f := h \circ g$.

\textbf{Class-incremental learning (CIL)} is a setting in continual learning that considers learning a sequence of $N$ tasks 
$\{\mathcal{T}_1, \mathcal{T}_2, \dots, \mathcal{T}_N\}$ over $(N-1)$ incremental steps.
Let $T$ denote the number of tasks learned so far.
Each task $\mathcal{T}_t$ $(1 \leq t \leq T)$ is associated with a training dataset 
$\mathcal{D}_t = \{(\boldsymbol{x}_t^i, y_t^i)\}_{i=1}^{n_t}$, 
where $n_t$ is the number of samples in task $\mathcal{T}_t$, 
$\boldsymbol{x}_t^i \in \mathcal{X}$ is an input, and 
$y_t^i \in \mathcal{Y}_t = \{1, 2, \dots, C_t\}$ is its class label, with $C_t$ denoting the number of classes in task $\mathcal{T}_t$.
The class sets are mutually disjoint, i.e., $\mathcal{Y}_t \cap \mathcal{Y}_{t'} = \emptyset$, $\forall t \neq t'$, 
and $\bigcup_{t=1}^{T} \mathcal{Y}_t = \mathcal{Y}$, where $\mathcal{Y} = \{1, 2, \dots, C\}$ and $C = \sum_{t=1}^{T} C_t$ denotes the total number of classes observed up to task $T$.
The goal of CIL is to learn a single predictive model 
$f : \mathcal{X} \rightarrow \mathbb{R}^C$ 
that can correctly predict the label $y \in \mathcal{Y}$ for any test sample $\boldsymbol{x}$ drawn from the classes observed up to task $T$.


\textbf{Kolmogorov--Arnold Networks (KAN).}
KAN~\cite{KAN} is a neural network with a unique architecture that replaces shared weights and fixed activation functions in traditional multi-layer perceptrons (MLPs)~\cite{MLP} with trainable activation functions. 
This imparts KANs with the property of local neuroplasticity.
As a result, learning a new task primarily impacts the localized regions of the network activated by the training data, while preserving the integrity of distant regions.
This property makes KANs naturally effective in mitigating catastrophic forgetting~\cite{KAC}. 

KANs are founded on the Kolmogorov--Arnold representation theorem (KAT)~\cite{KAT}, which states that any multivariate continuous function $f(\boldsymbol{x})$ defined on a bounded domain can be represented as a finite composition of univariate continuous functions combined through addition.
Formally, the theorem can be written as:
\begin{equation}
    f(\boldsymbol{x}) = f(x_1, x_2, \ldots, x_n) = \sum_{q=1}^{2n+1} \Phi_q \left( \sum_{p=1}^n \phi_{q,p}(x_p) \right),
\end{equation}
where $\Phi_q$ and $\phi_{q,p}$ are univariate continuous functions.
In KAN, the functions $\phi_{q,p}$ and $\Phi_q$ are parameterized using learnable coefficients over predefined basis functions, typically implemented as B-spline bases~\cite{Bspline}.

\textbf{Radial basis function Kolmogorov--Arnold Networks (RBF-KAN).}
Several recent studies~\cite{RBFKAN, RBFKAN2} have shown that the B-spline basis functions used in the original KAN~\cite{KAN} can be replaced by radial basis functions (RBF)~\cite{RBF} to more effectively represent high-dimensional data, thus reducing the curse of dimensionality (COD).
Furthermore,~\cite{KAC} shows that the limited fitting capacity of B-spline functions on high-dimensional data can sometimes further increase forgetting when introduced into CIL.
Therefore, we study RBF-KAN in the OOD-CL setting, in place of the original KAN.

Specifically, an RBF-KAN layer can be expressed as:
\begin{equation} \label{eq:RBFKAN}
f(\boldsymbol{x}) = \sum_{p=1}^n \Phi_p \left( \sum_{i=1}^G \omega_{p,i} \, \phi\bigl(\|x_p - c_i\|\bigr) \right),
\end{equation}
where $c_i \in \mathbb{R}$ denotes a set of center points evenly distributed within a specific range, and $G \in \mathbb{N}^{+}$ is the total number of such centers.
The function $\phi(\cdot): \mathbb{R} \rightarrow \mathbb{R}^{+}$ is an RBF kernel serving as the basis function, whose value depends only on the distance between the input $x_p$ and the center $c_i$.
The coefficients $\omega_{p,i}$ are learnable weights associated with each basis function.

A common choice for $\phi$ is the Gaussian kernel~\cite{RBFKAN}, defined as:
\begin{equation} \label{eq:RBF}
\phi\bigl(\|x_p - c_i\|\bigr) = \exp\left(-\frac{(x_p - c_i)^2}{2\sigma_i^2}\right),
\end{equation}
where $\sigma_i > 0$ controls the width (or spread) of the kernel.
This formulation imparts locality to the network: the response is highest near the center $c_i$ and decays rapidly with distance. 
Such localized behavior enables RBF-KAN to effectively capture local structures and adapt efficiently to local changes in the data~\cite{RBFKAN2}.

\textbf{RBF-KAN as a classifier.}
Following~\cite{KAC}, we can construct an RBF-expanded classifier inspired by KAN operating on the normalized embedding $g(\boldsymbol{x}) =  \boldsymbol{z} = [z_1, \dots, z_D] \in \mathbb{R}^D$, where $D$ is the embedding dimension.
If we define $\Phi_p$ in \cref{eq:RBFKAN} as a learnable linear weight for each dimension and consolidate these weights with those corresponding to  $\phi$ into a single learnable weight matrix $W \in \mathbb{R}^{C \times(D \times G)}$, the KAN-based classifier $h_{\mathrm{KAN}}$ can be expressed as:
\begin{equation} \label{eq:hkan}
h_{\mathrm{KAN}}(g(\boldsymbol{x})) = W \cdot \operatorname{vec}\!\left(\Phi(g(\boldsymbol{x}))\right),
\end{equation} 
where $\Phi(g(\boldsymbol{x})) \in \mathbb{R}^{D \times G}$ is the RBF expansion matrix, which can be calculated as:
$$
\begin{bmatrix}
\phi(\|z_1 - c_1\|) & \phi(\|z_1 - c_2\|) & \cdots & \phi(\|z_1 - c_G\|) \\
\phi(\|z_2 - c_1\|) & \phi(\|z_2 - c_2\|) & \cdots & \phi(\|z_2 - c_G\|) \\
\vdots & \vdots & \ddots & \vdots \\
\phi(\|z_D - c_1\|) & \phi(\|z_D - c_2\|) & \cdots & \phi(\|z_D - c_G\|)
\end{bmatrix},
$$
and $\operatorname{vec}(\cdot)$ flattens the $D \times G$ matrix into a vector for compatibility with $W$.


\textbf{Prompt-Based CIL Models.}
Due to recent advancements in pre-trained models and large language models (LLMs), prompt learning methods have been widely recognized as the state-of-the-art~\cite{CODA_prompt,CPrompt, Bisecle, L2P, Dual_prompt} in the challenging CIL setting, owing to their scalability and ability to mitigate catastrophic forgetting without buffering past examples~\cite{KAC}. 
These methods keep the pre-trained backbone $g$ (typically a sequence model such as ViT~\cite{vit}) frozen for improved knowledge transfer and attach a small set of learnable parameters, called prompts, to instruct the classifier $h$ in learning tasks that arrive sequentially.
Thus, each task $\mathcal{T}_t$ is associated with a set of prompt parameters 
$\boldsymbol{p}_t = \{\boldsymbol{p}_t^i\}_{i=1}^{M}$, where $\boldsymbol{p}_t^i \in \mathbb{R}^{L_{\boldsymbol{p}} \times D}$, 
with $L_{\boldsymbol{p}}$ denoting the prompt length and $M$ the number of prompts.
These prompts are prepended to the embedding features to perform classification.   
In addition, each $\boldsymbol{p}_t$ is associated with a corresponding set of keys 
$\boldsymbol{k}_t = \{\boldsymbol{k}_t^i\}_{i=1}^{M}$, where $\boldsymbol{k}_t^i \in \mathbb{R}^{D}$; 
these are also learnable parameters designed to capture representative features of each task.
Then, to select the best matching prompts for a given sample, a query--key matching mechanism~\cite{L2P} is employed.
Commonly, the backbone $g$ is used as a frozen feature extractor to obtain a query representation $q(\boldsymbol{x}) = g(\boldsymbol{x})[0]$, corresponding to the [CLS] token~\cite{vit}, and cosine similarity $\gamma$ is employed as the matching score. 
During training, for samples arriving in each task, a matching loss is then employed to pull the query features closer to the corresponding task-specific keys:
\begin{equation} \label{eq:Lmatch}
\mathcal{L}_{\text{match}}(\boldsymbol{x}, \boldsymbol{k}_t) = \gamma\big(q(\boldsymbol{x}), \boldsymbol{k}_t\big), \quad \boldsymbol{x} \in \mathcal{D}_t.
\end{equation} 



\section{Methodology} \label{sec:method}

\subsection{Problem Formulation}
\textbf{Out-of-distribution detection in continual learning (OOD-CL).}
We define OOD-CL as the CIL setting with the OOD detection capability.
At any time, the learned CIL model is required to (i) correctly classify test samples belonging to the classes observed in the $T$ tasks learned so far, and (ii) detect samples that do not belong to any of the learned classes so far, i.e., OOD samples.
Importantly, OOD samples may correspond to classes that will be introduced in future tasks. 
As learning progresses, these samples eventually become in-distribution, requiring the model to correctly incorporate them without forgetting previously learned knowledge.

We adopt an evaluation protocol based on standard CIL~\cite{CIL1st}, following prior works~\cite{proKT,OODCILunsuper, MORE1}, where OOD detection is evaluated before each incremental learning step.
Formally, let $\{\mathcal{S}_1, \dots, \mathcal{S}_N\}$ denote the test datasets for each task.
At the beginning of the learning step for the new task $\mathcal{T}_{T+1}$, the test data from previously learned tasks $\{\mathcal{S}_t\}_{t=1}^{T}$ are treated as ID, while the test data from the incoming task $\mathcal{S}_{T+1}$ are treated as OOD, where OOD detection is performed a total of $(N - 1)$ times while continually learning the sequence of $N$ tasks $\{\mathcal{T}_1, \dots, \mathcal{T}_N\}$.

Note that OOD-CL differs from the traditional static OOD detection setting, which has access to the full ID data at once.
In contrast, in OOD-CL, ID data arrive incrementally across tasks, and the training data available for OOD detection prior to learning $\mathcal{T}_{T+1}$ are restricted to the training dataset $\mathcal{D}_T$.
Moreover, unlike other methods~\cite{H2ST, MORE1, MORE}, we do not assume any form of rehearsal buffer.

\begin{wrapfigure}{r}{0.45\linewidth}
\vspace{-10pt}
\centering
\begin{subfigure}{\linewidth}
    \centering
    \includegraphics[width=\linewidth]{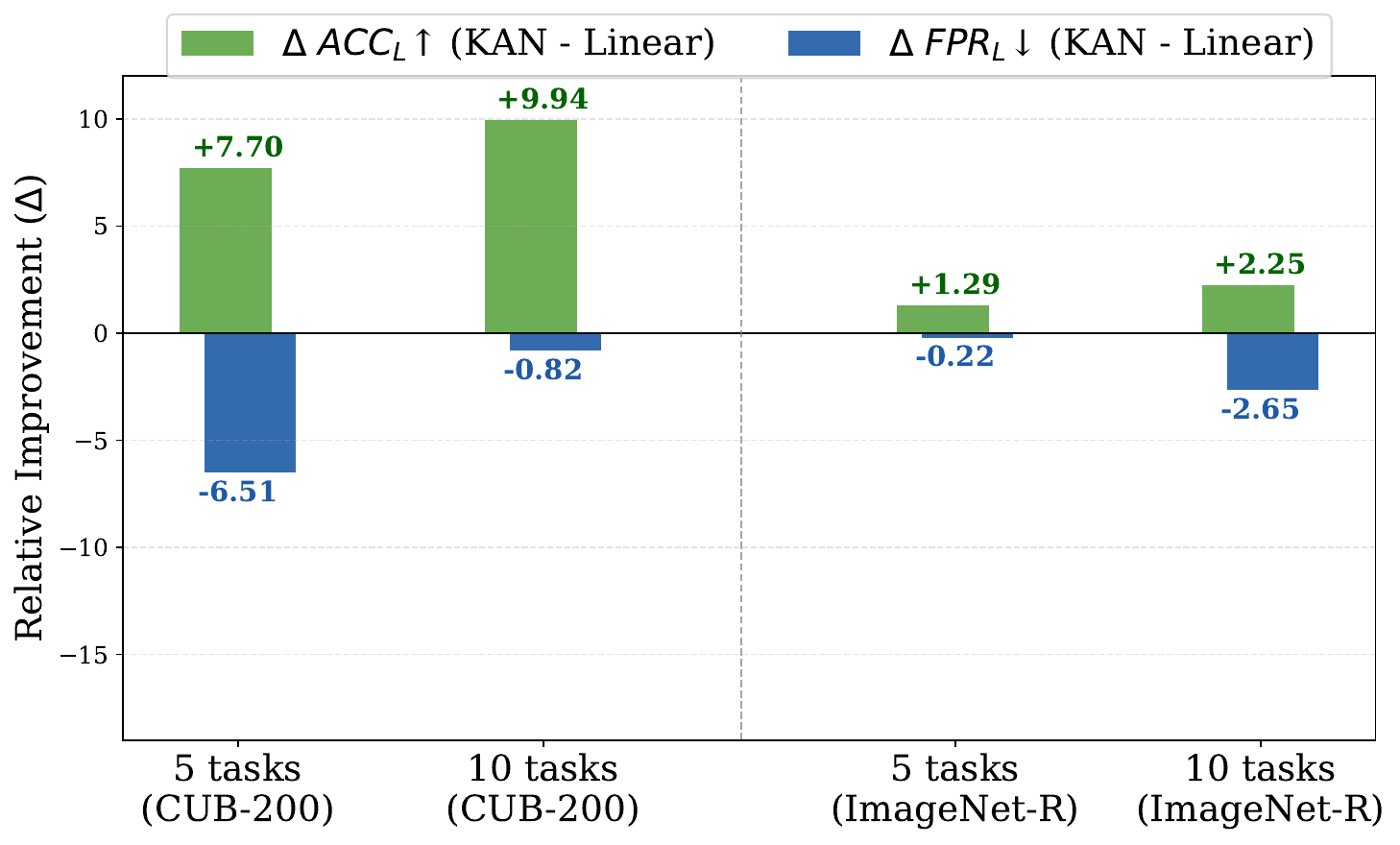}
    \label{fig:KvsL}
\end{subfigure}
\begin{subfigure}{\linewidth}
    \centering
    \includegraphics[width=\linewidth]{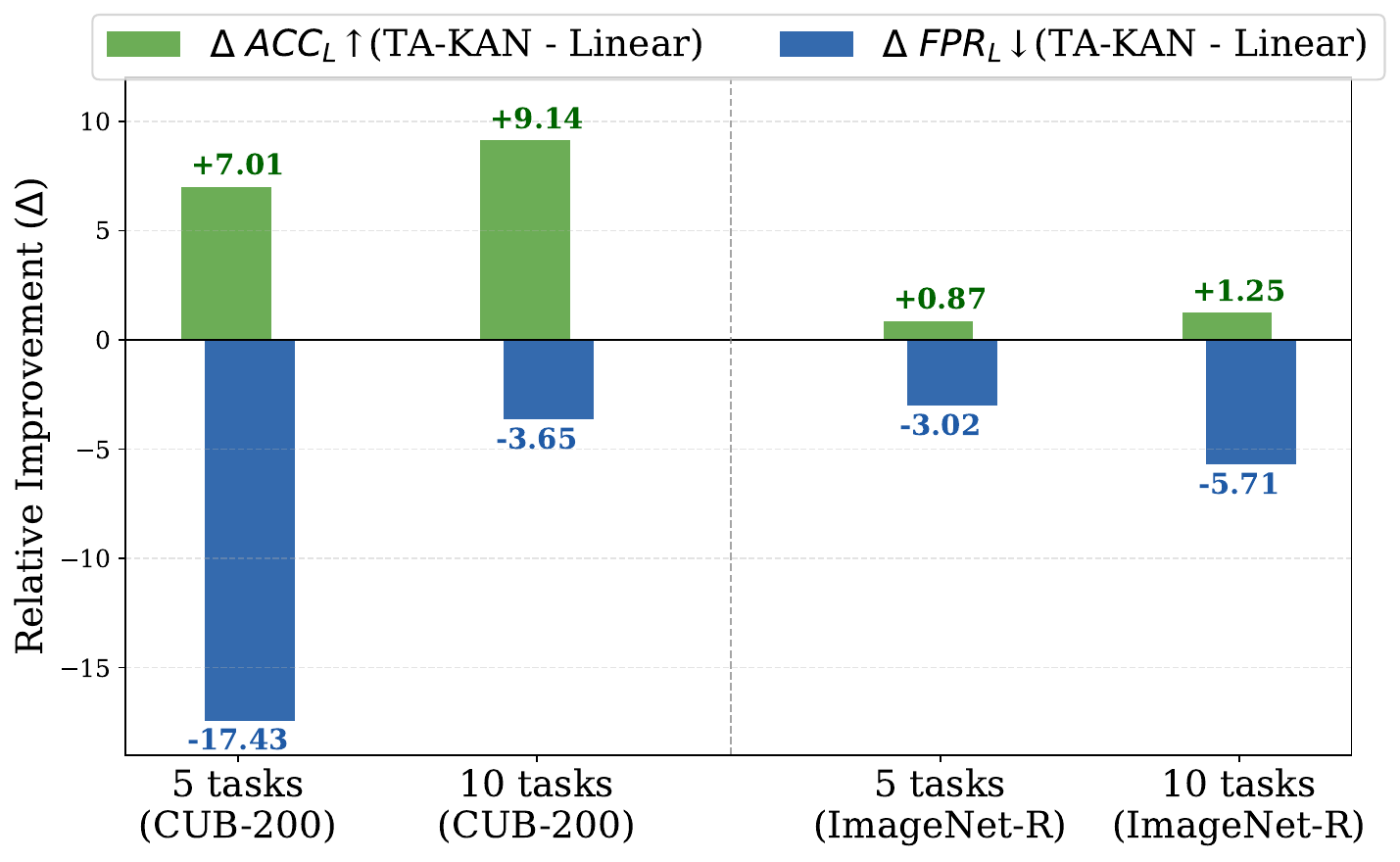}
    \label{fig:aKvsL}
\end{subfigure}
\vspace{-25pt}
  \caption{Relative improvement ($\Delta$) of the KAN head over a linear head in a SOTA CIL model. The improvement in OOD detection is limited compared to ID performance. The proposed TA-KAN significantly improves OOD performance while maintaining comparable ID performance.}
\label{fig:KANvsLinear}
\vspace{-15pt}
\end{wrapfigure}

\subsection{Motivation}
Recent continual learning methods, including prompt-based CIL approaches~\cite{CIL_KD,CPrompt,CODA_prompt,Dual_prompt}, typically employ an expandable linear classifier $h$ on top of the frozen backbone to accommodate newly encountered classes. 
Even though they freeze encoder weights, forgetting can still occur in downstream components, suggesting that careful design of the classifier head is important~\cite{KAC}.
Recent work~\cite{KAC} has shown that the conventional linear classifier can be effectively replaced by a KAN-based classifier, owing to its inherent local plasticity, which helps mitigate catastrophic forgetting. 
Building on this, we adopt a KAN-based classifier in the OOD-CL setting, aiming to improve both ID classification and OOD detection, on a generic prompt-based CIL model.

In the open-world setting, we observe that KAN also improves OOD detection performance alongside its gains in ID classification (\cref{fig:KANvsLinear}). 
However, despite its strong resistance to catastrophic forgetting, the improvement in OOD detection over a linear classifier remains limited relative to the gains observed in ID performance, where we believe something degrades the model’s ability to reliably distinguish OOD samples.
To investigate this, we analyze a trained CIL model after learning $N-1$ tasks and evaluate its OOD detection performance with respect to each task from $1$ to $N-1$ as ID in turn.
As shown in \cref{fig:bias_plot}, OOD detection performance is considerably lower when older tasks are treated as ID compared to when recently learned tasks are considered as ID. 
This observation indicates that OOD discrimination remains biased toward classes from newly learned tasks, which in turn degrades the model’s ability to reliably distinguish OOD samples.
A natural way to address this bias is to adapt the model behavior across tasks. 
However, explicitly maintaining task-specific models or parameters violates the CIL setting~\cite{parsimony_CIL, CIL1st}, which requires a single shared model without access to task identity at inference time. 
Therefore, we propose task-adaptive KAN (\textit{TA-KAN}), which preserves parsimony while enabling task-dependent behavior.

\subsubsection{Task Adaptive KAN (TA-KAN)}
We employ a CIL model with a KAN-based classifier head $h$ as defined in \cref{eq:hkan}. 
It utilizes an RBF-KAN layer consisting of $G$ Gaussian radial basis functions uniformly centered at $c = [c_1, c_2, \cdots, c_G]$ to approximate the activation function for each hidden feature dimension. 
Typically, the width of each RBF, $\sigma$, is set proportional to the step size $s = \frac{b-a}{G-1}$, which yields uniform coverage over the input range $[a,b] \subset \mathbb{R}$. 
However, we observe that the RBF width directly controls the locality of the response: narrower widths produce sharper, more localized activations, while wider widths yield smoother, more generalized responses. 
Therefore, in our TA-KAN, we adaptively assign RBF widths across the activation region, using narrower RBFs in regions relevant to old classes compared to those for relatively newer tasks.
This encourages older classes to form sharper and more distinct activation regions, making them easier to separate from nearby OOD samples.

\begin{figure}[t]
\vspace{-10pt}
    \centering
    \begin{subfigure}[t]{0.25\linewidth}
        \centering
        \includegraphics[height=0.15\textheight,keepaspectratio]{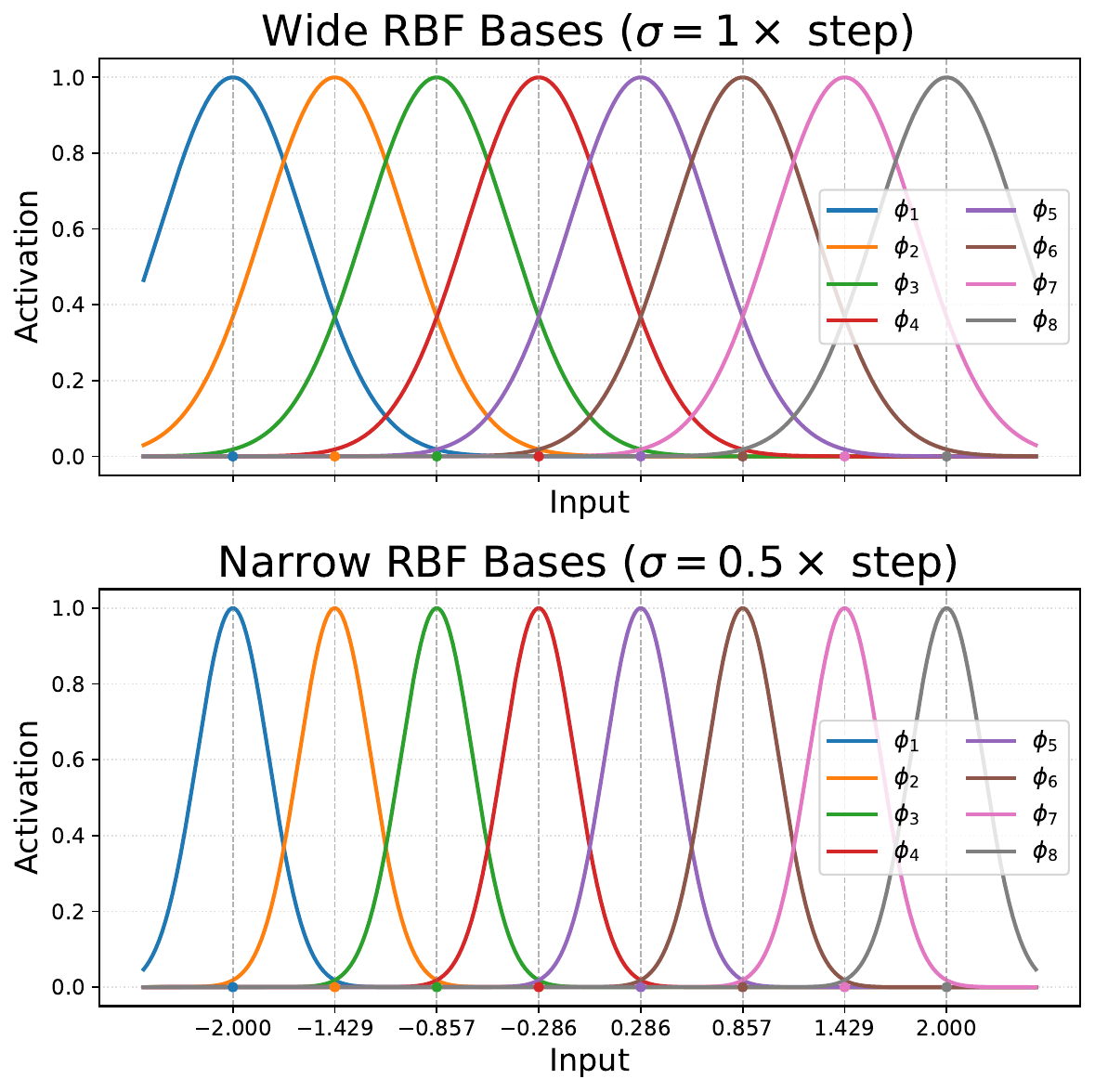}
        \caption{RBF bases}
        \label{fig:rbf_bases}
    \end{subfigure}
    \hfill
    \begin{subfigure}[t]{0.35\linewidth}
        \centering
        \includegraphics[height=0.15\textheight,keepaspectratio]{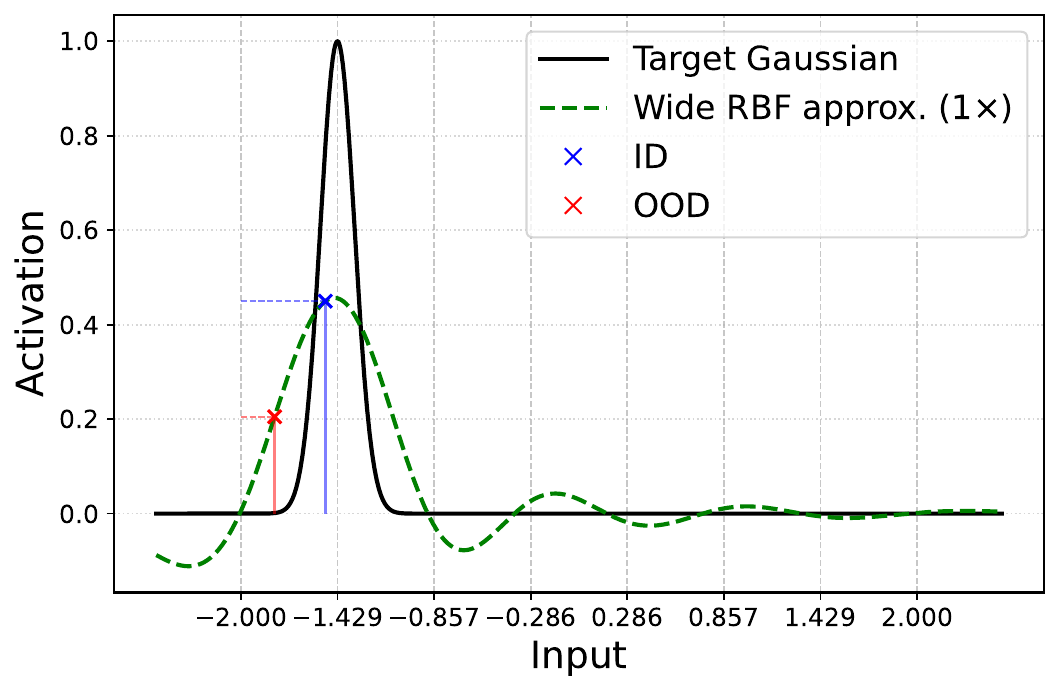}
        \caption{Wide RBF approx.}
        \label{fig:rbf_wide_marked}
    \end{subfigure}
    \hfill
    \begin{subfigure}[t]{0.35\linewidth}
        \centering
        \includegraphics[height=0.15\textheight,keepaspectratio]{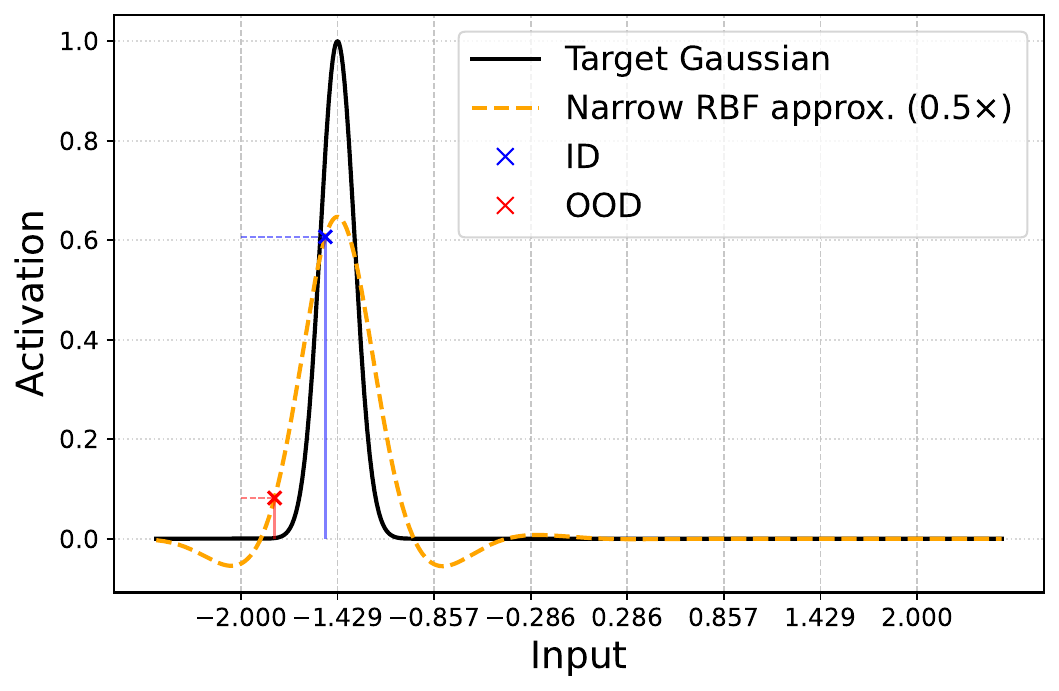}
        \caption{Narrow RBF approx.}
        \label{fig:rbf_narrow_marked}
    \end{subfigure}
\caption{Comparison of RBF approximations using wide and narrow basis functions. 
(a) RBF bases with different widths, $1\times s$ (top) and $0.5\times s$ (bottom). 
(b) Approximation of a target ID activation using wide bases yields a coarse approximation and produces similar responses for both ID and nearby OOD samples. 
(c) Approximation using narrower bases with similar weights, results a more accurate reconstruction and better captures localized responses.}
    \label{fig:RBF_approx}
    \vspace{-15pt}
\end{figure}

To illustrate this effect, we provide a toy example in \cref{fig:RBF_approx}, where an ID region is characterized by a narrow Gaussian activation ($\sigma = 0.25 \times s$), and an OOD sample is located close to this region. 
Narrower RBF bases can better approximate the sharp local structure of the ID region than wider RBFs, resulting in more distinct activations and reduced overlap between ID and OOD responses while using similar weights in both cases.

However, naively adopting a uniformly narrower RBF width across the entire activation region can make the model sensitive to intra-class variations and reduce its generalization ability, potentially degrading the ID classification performance (cf. Sec.~\ref{sec:ablation}).
Instead, we adapt the RBF width to focus the model’s capacity on harder, older tasks while maintaining sufficient generalization for relatively easier, newly learned tasks.


\textbf{Identifying the activation ranges of interest.}
A key challenge in implementing the adaptive RBF network is how to identify the activation range of interest for a given sample—particularly for OOD inputs—in order to select its desired RBF width. 
To address this, we repurpose how the prompt based CIL model employs the similarity-based estimation for a sample to select its most relevant prompts \cite{L2P, Dual_prompt, CPrompt, KAC} or to reweight prompts \cite{CODA_prompt, Bisecle} without introducing additional complexity or parameters specifically for OOD detection.
The model incorporates query-key matching to identify task specific prompts, which we leverage to estimate the task affinity of a sample and localize its corresponding activation region of interest.
Specifically, given the query representation $q(\boldsymbol{x})$ and similarity measure $\gamma(\cdot,\cdot)$, we estimate the task affinity by selecting the task whose key exhibits the highest alignment with the query:
\[
\hat{t}(\boldsymbol{x}) = 
\underset{t \in \{1,\dots,T\}}{\arg\max} 
\max_{i \in \{1,\dots,M\}} 
\gamma\big(q(\boldsymbol{x}), \boldsymbol{k}_t^i\big).
\]
For ID samples, the inferred $\hat{t}(\boldsymbol{x})$ corresponds to their true task, whereas for OOD samples, it reveals the ID task whose feature manifold they most closely resemble. 
Importantly, for OOD samples, this mechanism allows us to identify the region of the ID feature space where the model is most likely to exhibit ambiguity, i.e., the task whose decision boundaries are most relevant—and potentially most vulnerable—for distinguishing $\boldsymbol{x}$ from ID data.
We therefore use $\hat{t}(\boldsymbol{x})$ to guide the adaptive RBF width to improve separability in that region.

\textbf{TA-KAN classifier.} 
In our proposed TA-KAN classifier, the activation range of each task is approximated using RBF bases of unique width assigned to it.
To achieve this, we parameterize the Gaussian basis using a task-specific scaling factor. 
Specifically, let $\alpha_t$ denote the width factor associated with task $\mathcal{T}_t$. 
At inference, the width is selected based on the inferred task affinity $\hat{t}(\boldsymbol{x})$, i.e., $\alpha_{\hat{t}(\boldsymbol{x})}$.
Accordingly, the Gaussian basis in \cref{eq:RBF} is defined as
\begin{equation}
\phi\left(\left\|z_p-c_i\right\| ; \alpha_{\hat{t}(\boldsymbol{x})}\right)=
\exp\!\left(
-\frac{(z_p-c_i)^2}{2\left(\alpha_{\hat{t}(\boldsymbol{x})} s\right)^2}
\right).
\end{equation}
\looseness=-1
This yields a task-conditioned RBF expansion matrix $\Phi\!\left(g(\boldsymbol{x}); \alpha_{\hat{t}(\boldsymbol{x})}\right)$, and the classifier is expressed as 
\begin{equation}
h_{\mathrm{A\mbox{-}KAN}}\!\left(g(\boldsymbol{x}); \alpha_{\hat{t}(\boldsymbol{x})}\right)
=
W \cdot \operatorname{vec}\!\left(
\Phi\!\left(g(\boldsymbol{x}); \alpha_{\hat{t}(\boldsymbol{x})}\right)
\right).
\end{equation}
Importantly, the classifier shares a single RBF layer across all tasks, thereby preserving model parsimony. 
Instead, task adaptivity is introduced solely through a sample-dependent basis expansion via adaptive width selection.\footnote{In \cref{eq:alphat}, the width factors $\{\alpha_t\}$ are deterministically predefined using a simple schedule parameterized by $\alpha_{\min}$ (with $\alpha_{\max}=1$), and thus introduce no additional learnable parameters.}
\\Furthermore, we define the task-specific width factors $\{\alpha_t\}_{t=1}^N$ as
\begin{equation} \label{eq:alphat}
\alpha_t = \alpha_{\min} + \frac{t-1}{N-1}(\alpha_{\max} - \alpha_{\min}),
\end{equation}
where $0 < \alpha_{\min} \le \alpha_t \le \alpha_{\max} \le 1$.
Note that $\alpha_t$ increases monotonically with task recency, such that earlier tasks are assigned smaller widths, resulting in sharper and more localized activations, while more recent tasks retain relatively wider responses to preserve generalization.

During training, we use the ground-truth task identity $t$ to select the corresponding width factor $\alpha_t$, and optimize the model end-to-end.
The overall objective is given by
\begin{equation}
\mathcal{L} = \mathcal{L}_{\text{CE}}\left(h_{\mathrm{A\mbox{-}KAN}}\!\left(g(\boldsymbol{x}); \alpha_t\right), y\right)+\lambda \mathcal{L}_{\text {match }}\left(\boldsymbol{x}, \boldsymbol{k}_t\right), \quad \boldsymbol{x} \in \mathcal{D}_t,
\end{equation}
where $\mathcal{L}_{\text{CE}}$ is the cross-entropy loss, $\mathcal{L}_{\text{match}}$ is the matching loss defined in \cref{eq:Lmatch}, and $\lambda$ is a balancing coefficient.




\subsubsection{Geometry-Based Guidance for OOD-CL}
Furthermore, since the KAN-based CIL model is optimized for improving generalizability to mitigate catastrophic forgetting, this can also induce lower confidence for ID data compared to the static setting \cite{OODCILunsuper}.
Since standard OOD scoring functions (e.g., MSP~\cite{hendrycks_2017MSP}, MaxLogit~\cite{MaxLogit}) rely solely on the confidence associated with the classifier, this can lead to increased overlap between ID and OOD scores, reducing their separability.
Moreover, these methods are known to suffer from the overconfidence issue~\cite{hendrycks_2017MSP,NNGuide}, which further limits their reliability.
To mitigate the overconfidence associated with such classifier-derived OOD scores, recent OOD methods turn to non-classifier signals, such as distance- and geometry-based measures in the feature space (KNN~\cite{KNN_OOD}, Mahalanobis~\cite{lee_2018_mahalanobis}) or combine both types of information (NNGuide~\cite{NNGuide}, LID~\cite{SupLID}, VIM~\cite{VIM}).
However, these geometry-based approaches typically require access to the full training in-distribution, making them inapplicable in the CIL setting.
In this section, we propose a novel geometry-based guidance mechanism tailored for OOD-CL.
Specifically, instead of modeling the geometry of the data manifold in the feature space, we exploit the prompt space, which captures higher-level task semantics beyond the lower-level embedding representations.
\\ Following~\cite{NNGuide}, a geometry-based guidance can be introduced to $S$ as
\begin{equation} \label{eq:sguide}
S_{\mathrm{Guided}}(\boldsymbol{x})=S(\boldsymbol{x}) \cdot \Gamma(\boldsymbol{x}),
\end{equation}
where $\Gamma(\boldsymbol{x})$ represents a distance-based guidance score.
We choose $\Gamma(\boldsymbol{x})$ as the average similarity to the $k$-nearest neighbors in the ID prompt space,
\begin{equation} \label{eq:Dknn}
\Gamma(\boldsymbol{x}) = \frac{1}{k} \sum_{\boldsymbol{p} \in \mathcal{N}_k(\boldsymbol{x})} 
\operatorname{sim}\big(q(\boldsymbol{x}), \boldsymbol{p}\big)
\end{equation}
where $\mathcal{N}_k(\boldsymbol{x})$ denotes the set of $k$ nearest prompts to sample $\boldsymbol{x}$.
To determine the set of prompts for a given input, we leverage the learned key representations.
A natural choice for $k$ is either $M$, the number of prompts allocated per task, or $1$, corresponding to selecting the single most similar prompt. 
In practice, we find that the latter yields slightly better performance.
Accordingly, we define the guidance score as the maximum similarity across all keys:
\begin{equation} \label{eq:Dmax}
\Gamma(\boldsymbol{x}) = \max_{t \in \{1,\dots,N\}, i \in \{1,\dots,M\}} 
\gamma\big(q(\boldsymbol{x}), \boldsymbol{k}_t^i\big).
\end{equation}
Unless otherwise stated, we adopt MSP \cite{hendrycks_2017MSP} as the default classifier-based confidence score $S(\boldsymbol{x}) \in [0,1]$, owing to its generality.
Since cosine similarity $\gamma(\cdot,\cdot)\in [-1,1]$, we further rescale $\Gamma(\boldsymbol{x})$ to $[0,1]$ as
$\tilde{\Gamma}(\boldsymbol{x}) = \frac{\Gamma(\boldsymbol{x}) + 1}{2},$
so that the guided score remains bounded, i.e., $S_{\mathrm{Guided}}(\boldsymbol{x})=S(\boldsymbol{x}) \cdot \tilde{\Gamma}(\boldsymbol{x}) \in [0,1]$.

In the proposed $S_{\mathrm{Guided}}$ score, $S$ mostly captures intra-task discrimination, while $\Gamma$ captures inter-task structure to guide OOD detection.
Importantly, $\Gamma$ provides complementary guidance that helps further reduce the bias and overconfidence associated with classifier-based confidence scores.

\textbf{Two-level manifold constraint for OOD-CL. }
Moreover, the proposed $S_{\mathrm{Guided}}$ can be interpreted as enforcing a two-level manifold constraint for continual OOD detection. 
To illustrate this, let $\boldsymbol{w}_c \in \mathbb{R}^D$ denote the weight vector corresponding to class $c$ in the classifier, and consider a simplified setting where $M = 1$.
Then, the proposed score can be written as
\begin{equation}
S_{\mathrm{Guided}} := 
\underbrace{\max_{c \in \{1, \ldots, C\}} \boldsymbol{w}_c^{\top} \boldsymbol{z}}_{\text{class alignment}} 
\;\cdot\;
\underbrace{\max_{t \in \{1, \ldots, N\}} \boldsymbol{k}_t^{\top} \boldsymbol{z}}_{\text{task alignment}}.
\end{equation}
In fact, each weight $\boldsymbol{w}_c$ can be viewed as a learned class prototype, while each key $\boldsymbol{k}_t$ serves as a task-level prototype, since under $\mathcal{L}_{\text{match}}$ it is optimized to align with features of its corresponding task.
The first term measures the alignment between the feature representation and learned class prototypes, while the second term captures alignment with learned task-level prototypes.
Consequently, a sample is considered in-distribution only if it aligns well with both a class manifold and a task manifold, thereby improving OOD robustness by suppressing samples that fail to align at either level.


\section{Evaluation}\label{sec:evaluation}

\subsection{Experimental Settings} \label{sec:ex_settings}

\textbf{Datasets and Tasks.}
We evaluate the proposed OOD-CL method across diverse CIL settings using three widely adopted benchmarks~\cite{CODA_prompt,KAC,Dual_prompt}: ImageNet-R~\cite{ImageNet-R}, CUB-200~\cite{CUB} and CIFAR-100~\cite{cifar}. 
For CUB-200 and CIFAR-100, we consider 5-task (40 classes per task), 10-task (20 classes per task), and 20-task (10 classes per task) settings. 
For ImageNet-R, we adopt 5-task, 10-task, 20-task and 40-task settings.
All experiments are conducted in a non-exemplar setting, where no samples from previous tasks are stored for future training. 
Following standard CIL, results are averaged over multiple random seeds with randomly shuffled task orders.

\textbf{Evaluation Metrics.}
We use widely adopted metrics for OOD-CL~\cite{proKT, MORE}: the area under the receiver operating characteristic curve ($AUC_L$), false positive rate at 95\% true positive rate ($FPR_L$), and average precision for ID and OOD samples ($AUP_{L_\mathrm{In}}$ and $AUP_{L_\mathrm{Out}}$, respectively), measured after learning the last task and averaged over all seen tasks. These metrics provide a comprehensive measure of OOD detection performance across all tasks.
To evaluate ID classification performance, we adopt the average final accuracy ($ACC_L$),—i.e., the accuracy after learning the last task, averaged over all seen tasks, which is a widely used metric in CIL, capturing both plasticity and forgetting~\cite{CODA_prompt, CPrompt}.
We also report the average incremental performance over all learned tasks for both OOD detection ($AUC_A$, $FPR_A$) and ID classification ($ACC_A$).

\textbf{Baselines and configuration.}
We evaluate our method using five OOD detection methods: MSP~\cite{hendrycks_2017MSP}, MaxLogit~\cite{MaxLogit}, Energy~\cite{EnergyOE}, Entropy~\cite{MetaOOD}, KL~\cite{MaxLogit} and three OOD-CL methods: MORE~\cite{MORE}, MSP-BC-CE~\cite{OODCILunsuper} and Pro-KT~\cite{proKT}.
To evaluate the effectiveness of our method, we adopt CODAPrompt~\cite{CODA_prompt} as the backbone prompt model, following~\cite{KAC}, while noting that our approach generalizes to other prompt-based models (see \cref{app:ablation_prompt}).
CODAPrompt employs a linear classifier as $h$ (\textit{Linear-Head}). 
Following~\cite{KAC}, we first replace this classifier with a KAN (\textit{KAN-Head})~\cite{KAC}, and then with our proposed TA-KAN classifier (\textit{TA-KAN}).
We additionally consider two canonical classifier alternatives for mitigating classifier bias: a cosine-normalized classifier (\textit{Cosine-Head})~\cite{coshead} and a nearest-class-mean classifier (\textit{NCM Readout})~\cite{CIL1st}.
All classifier variants are evaluated within the same prompt-learning framework under otherwise consistent configurations to ensure a fair comparison of OOD-CL performance.
See \cref{app:configuration} for further details.

\subsection{Experimental Results}

\begin{table}[t!]
\centering
\caption{Results on CUB-200~\cite{CUB} dataset under different task settings. Lower FPR is better; higher AUC, AUP and ACC are better.}
\small
\renewcommand{\arraystretch}{1.0}
\setlength{\tabcolsep}{0.5pt}
\resizebox{\textwidth}{!}{
\begin{tabular}{lccccc|ccccc|ccccc}
\toprule
\multirow{2}{*}{\textbf{Method}}
& \multicolumn{5}{c}{\textbf{5 Tasks}}
& \multicolumn{5}{c}{\textbf{10 Tasks}}
& \multicolumn{5}{c}{\textbf{20 Tasks}} \\
\cmidrule(lr){2-6} \cmidrule(lr){7-11} \cmidrule(lr){12-16}
& $FPR_L$ & $AUC_L$ & $AUP_{L_\mathrm{In}}$ & $AUP_{L_\mathrm{Out}}$ & $ACC_L$
& $FPR_L$ & $AUC_L$ & $AUP_{L_\mathrm{In}}$ & $AUP_{L_\mathrm{Out}}$ & $ACC_L$
& $FPR_L$ & $AUC_L$ & $AUP_{L_\mathrm{In}}$ & $AUP_{L_\mathrm{Out}}$ & $ACC_L$ \\
\midrule

Linear-Head~\cite{CODA_prompt}
& 66.16 & 79.30 & 77.08 & 80.43 & 73.85
& 71.93 & 77.41 & 74.78 & 79.64 & 68.35
& 67.91 & 79.56 & 77.44 & 81.57 & 61.69 \\

Cosine-Head~\cite{coshead}
& 60.67 & 84.13 & 81.20 & 85.24 & 80.08
& 69.40 & 80.30 & 76.11 & 82.12 & 73.17
& 63.19 & 79.14 & 78.11 & 80.14 & 62.24 \\

NCM Readout~\cite{CIL1st}
& 77.21 & 77.83 & 73.41 & 79.96 & 80.54
& 78.12 & 74.19 & 69.71 & 76.67 & 77.13
& 76.41 & 76.28 & 73.32 & 78.27 & \textbf{81.42} \\

KAN-Head (Baseline)~\cite{KAC}
& 59.65 & 83.48 & 80.51 & 84.27 & \textbf{81.55}
& 71.11 & 80.53 & 76.69 & 82.30 & \textbf{78.29}
& 59.69 & 81.86 & 81.18 & 82.93 & 75.24 \\

ProKT~\cite{proKT}
& 65.41 & 80.85 & 77.78	& 81.62 & 77.39
& 73.55 & 77.79 & 73.78	& 79.14 & 71.34
& 61.61 & 78.93 & 79.02	& 78.92 & 64.52 \\

MORE~\cite{MORE}
& 66.96 & 80.06 & 76.87	& 80.91 & 74.47
& 79.46 & 77.15 & 71.92	& 79.05 & 71.05
& 65.87 & 79.29 & 77.80	& 80.11 & 70.90 \\

MSP-BC-CE~\cite{OODCILunsuper}
& 58.01 & 84.60 & 81.17 & 85.34 & 81.55
& \textbf{67.88} & 80.66 & 77.66 & 82.36 & 78.29
& 60.65 & 80.20 & 79.28 & 81.77 & 75.24 \\
\midrule

TA-KAN
& \underline{52.11} & \underline{85.86} & \underline{83.49} & \underline{86.71} & 80.86
& 70.63 & \underline{81.55} & \underline{78.47} & \underline{83.52} & 77.49
& \underline{51.35} & \underline{84.42} & \underline{84.45} & \underline{84.79} & 75.70 \\

TA-KAN + Guidance
& \textbf{48.73} & \textbf{86.42} & \textbf{85.59} & \textbf{86.96} & \underline{80.86}
& \underline{68.28} & \textbf{81.99} & \textbf{80.25} & \textbf{83.87} & \underline{77.49}
& \textbf{50.48} & \textbf{84.77} & \textbf{84.85} & \textbf{85.10} & \underline{75.70} \\
\bottomrule
\end{tabular}
}
\label{tab:cub200_mainresults}
\vspace{-10pt}
\end{table}

\begin{table}[tbh!]
\centering
\caption{Results on ImageNet-R~\cite{ImageNet-R} dataset under different task settings. Lower FPR is better; higher AUC and ACC are better.}
\small
\renewcommand{\arraystretch}{1.0}
\setlength{\tabcolsep}{2pt}
\resizebox{0.99\textwidth}{!}{
\begin{tabular}{lccc|ccc|ccc|ccc}
\toprule
\multirow{2}{*}{\textbf{Method}}
& \multicolumn{3}{c}{\textbf{5 Tasks}}
& \multicolumn{3}{c}{\textbf{10 Tasks}}
& \multicolumn{3}{c}{\textbf{20 Tasks}}
& \multicolumn{3}{c}{\textbf{40 Tasks}} \\
\cmidrule(lr){2-4} \cmidrule(lr){5-7} \cmidrule(lr){8-10} \cmidrule(lr){11-13}
& $FPR_L$ & $AUC_L$ & $ACC_L$
& $FPR_L$ & $AUC_L$ & $ACC_L$
& $FPR_L$ & $AUC_L$ & $ACC_L$
& $FPR_L$ & $AUC_L$ & $ACC_L$ \\
\midrule
Linear-Head~\cite{CODA_prompt}        & 42.74 & 85.44 & 77.96 & 45.69 & 86.26 & 76.51 & 53.49 & 85.28 & 73.90 & 53.21 & 81.57 & 68.80 \\
Cosine-Head~\cite{coshead}                           & 41.87 & 84.62 & \textbf{80.73} & 42.32 & 86.73 & \textbf{79.22} & 45.64 & 85.50 & \textbf{77.91} & 50.17 & 83.27 & \underline{74.01} \\
NCM Readout~\cite{CIL1st}                        & 43.33 & 85.38 & 77.98 & 45.43 & 85.36 & 77.32 & 47.26 & 84.59 & 76.86 & 49.37 & 83.19 & \textbf{74.38} \\
KAN-Head (Baseline)~\cite{KAC}        & 42.52 & 84.81 & \underline{79.25} & 43.04 & 85.64 & \underline{78.76} & 44.00 & 85.40 & \underline{77.42} & 44.80 & 84.63 & 73.95 \\
ProKT~\cite{proKT}                    & 48.59 & 81.26 & 72.82 & 50.18 & 81.63 & 71.72 & 52.19 & 80.52 & 69.15 & 51.94 & 80.91 & 65.97 \\
MORE~\cite{MORE}                      & 45.82 & 80.74 & 78.63 & 49.21 & 80.12 & 74.29 & 53.67 & 80.31 & 71.82 & 59.61 & 79.32 & 70.45 \\
MSP-BC-CE~\cite{OODCILunsuper}        & 40.65 & 85.50 & {79.25} & 42.12 & \underline{86.84} & {78.76} & 44.88 & 85.78 & {77.42} & 47.21 & 84.81 & {73.95} \\
\midrule
TA-KAN                                & \underline{40.07} & \underline{85.77} & 78.83 & \underline{40.25} & 86.76 & 77.76 & \underline{42.57} & \underline{85.99} & 77.03 & \underline{44.16} & \textbf{86.48} & 72.38 \\
TA-KAN + Guidance                      & \textbf{39.72} & \textbf{85.92} & 78.83 & \textbf{39.98} & \textbf{86.87} & 77.76 & \textbf{42.61} & \textbf{86.10} & 77.03 & \textbf{44.03} & \underline{86.41} & 72.38 \\
\bottomrule
\end{tabular}
}
\label{tab:imagenetr_mainresults}
\vspace{-10pt}
\end{table}

We report results on the CUB-200 dataset in \cref{tab:cub200_mainresults} and ImageNet-R in \cref{tab:imagenetr_mainresults} across all task settings.
In addition to various classifier settings, we include SOTA OOD-CL methods, while comparisons with standard OOD detection methods are provided in \cref{app:ablation_posthoc}. 
For both datasets, replacing the linear classifier with a KAN-based classifier improves overall CIL performance, with particularly strong gains in ID classification accuracy. 
We then replace the standard KAN with the proposed TA-KAN classifier—designed specifically for OOD-CL—which significantly improves OOD detection performance in terms of both FPR and AUC, while preserving the ID performance gains achieved by KAN.
Further incorporating the proposed guidance into the TA-KAN scoring function yields additional improvements in OOD performance.
Notably, TA-KAN yields particularly large gains on CUB-200, a challenging benchmark for OOD-CL due to its fine-grained nature, reducing FPR by 12.57\% on average (up to 18.31\%) and improving AUC by 2.96\% on average (up to 3.55\%).
On ImageNet-R, even in scenarios where the KAN classifier degrades OOD detection performance relative to the linear baseline, TA-KAN recovers and improves performance.
Furthermore, our method significantly outperforms other OOD-CL methods, including MORE~\cite{MORE}, which relies on a rehearsal buffer. 
Additional datasets and analyses are provided in \cref{app:more_results}.

\textbf{Different OOD difficulty regimes.}
The different performance gains observed on CUB-200 and ImageNet-R can be understood through two difficulty regimes directly related to the locality-based design of TA-KAN: (i) fine-grained semantic ambiguity, where increased locality can improve ID--OOD separability, and (ii) high intra-class variation, where increased locality may reduce invariance.
CUB-200 represents the former, where classes exhibit subtle inter-class differences and are often close to near-OOD samples. 
Here, the increased locality induced by TA-KAN, particularly for older tasks, helps capture fine-grained deviations and improve ID--OOD separability, explaining its substantially larger gains over the standard KAN.
In contrast, ImageNet-R exhibits high intra-class diversity, with each class containing significant variations in style (e.g., cartoon, sketch, graffiti, origami).
While increased locality can improve discrimination in ambiguous regions, it can also reduce invariance to such intra-class variations, resulting in comparatively smaller gains.
This highlights an inherent trade-off between locality and generalizability: narrower RBFs improve fine-grained discrimination, whereas wider RBFs provide smoother responses that better accommodate large variations.
Overall, these results support the central design hypothesis of TA-KAN: adaptive locality improves fine-grained ID--OOD discrimination where needed while retaining sufficient generalizability to intra-class variation.

\subsection{Ablation Study and Analysis} \label{sec:ablation}
We evaluate the contribution of each component of our method in \cref{tab:ablation_summary}. 
\textbf{Importance of TA-KAN.}
To evaluate the effect of the proposed TA-KAN classifier, we first compare it with the standard KAN head. As shown, TA-KAN— even without guidance $\Gamma$—significantly improves OOD detection performance over the standard KAN, while preserving the ID performance gains over the linear head. 
This highlights the importance of a task-adaptive classifier architecture for OOD-CL, where adaptive control of locality improves the separation between ID and OOD samples, particularly for older tasks, thereby reducing bias and enhancing overall performance.
However, there exists an inherent trade-off between locality and generalizability in CIL classifiers. 
For example, using a KAN with uniformly narrower RBF bases (with $\sigma = \alpha_{\min}s$ and indicated as \textit{KAN-Head~$\alpha_{\min}$}) across all tasks—thereby increasing locality over the entire activation range—yields only marginal improvements over the standard KAN head, while degrading ID classification performance due to reduced generalization ability.
Moreover, simply varying $\alpha$ does not yield consistent improvements; see \cref{tab:alpha_fixed}.
To further control for classifier capacity, we compare against a parameter-matched \textit{MLP-Head}, since KAN and TA-KAN introduce additional parameters over a linear head due to the RBF expansion.
Its inferior OOD performance compared with both KAN and TA-KAN indicates that increased parameter count alone does not account for the observed gains, further supporting the benefit of the KAN architecture.
\begin{wraptable}{r}{0.6\linewidth}
\vspace{3pt}
\centering
\caption{Ablation study of different components. Results are averaged over all task settings on CUB-200 dataset.}
\small
\renewcommand{\arraystretch}{1.0}
\setlength{\tabcolsep}{1pt}
\resizebox{0.6\textwidth}{!}{
\begin{tabular}{lcccccc}
\toprule
\textbf{Method} & \textbf{$FPR_L$} & \textbf{$FPR_A$} & \textbf{$AUC_L$} & \textbf{$AUC_A$} & \textbf{$ACC_L$} & \textbf{$ACC_A$} \\
\midrule
MLP-Head &77.46	&64.09	&71.16	&77.72	&54.82	&70.49 \\
Linear-Head & 68.67 & 59.17 & 78.76 & 82.40 & 67.96 & 79.42 \\
Linear-Head+Guidance & 67.03	& 60.81	& 79.06	& 82.54	& 67.96	& 79.42 \\
\midrule
KAN-Head (Baseline) & 63.48 & 56.00 & 81.96 & 85.16 & \textbf{78.36} & \textbf{85.37} \\
KAN-Head $\alpha_{\min}$ & 60.12 & 54.28 & 82.27 & 85.80 & 76.11 & 84.52 \\
KAN-Head trainable $\sigma$ & 64.69 & 56.30 & 81.53 & 85.04 & 78.16 & 85.36 \\
\midrule
KAN-Head+Guidance & 58.46 & 54.13 & 82.75 & 85.83 & 78.36 & 85.37 \\
\midrule
TA-KAN Head & 58.03 & 55.16 & 83.94 & 85.64 & 78.02 & 85.24 \\
TA-KAN+Guidance (ours) & \textbf{55.83} & \textbf{52.18} & \textbf{84.39} & \textbf{86.18} & 78.02 & 85.24 \\
\bottomrule
\end{tabular}
}
\vspace{-15pt}
\label{tab:ablation_summary}
\end{wraptable}

\textbf{Comparison with trainable RBF bases.}
We also consider a variant of the KAN classifier where the RBF width is treated as a task-wise trainable parameter (\textit{KAN-Head~trainable~$\sigma$}), inspired by Free-KAN~\cite{freeKAN} architecture. 
However, this variant underperforms even the standard KAN. 
We attribute this to the fact that the learned RBF bases become overly adapted to ID task characteristics, limiting generalizability to unseen tasks.
This highlights the importance of the design principle in \cref{eq:alphat} for TA-KAN.

\textbf{Importance of Guidance.}
Applying the proposed guidance across classifier heads improves OOD detection, highlighting its role in complementing classifier confidence scores, which degrade in the absence of past data and are prone to overconfidence. 
The gains are particularly pronounced for KAN, where stronger task-age-dependent confidence degradation (\cref{fig:bias_plot}) creates greater ambiguity that the task-manifold guidance helps resolve. 
While TA-KAN mitigates much of this bias through adaptive locality, the guidance provides further complementary gains (cf. \cref{app:ablation_geometry}).

\begin{wraptable}{r}{0.4\linewidth}
\vspace{-13pt}
\caption{Effect of $\alpha$ on KAN and TA-KAN on the CUB-200 dataset. Lower FPR and higher AUC/ACC are better.}
\small
\setlength{\tabcolsep}{2pt}
\renewcommand{\arraystretch}{0.9}
\resizebox{0.4\textwidth}{!}{
\begin{tabular}{c l c c c}
\toprule

$\alpha$ & Method & $FPR_L$ & $AUC_L$ & $ACC_L$  \\
\midrule

\multirow{2}{*}{1}
& KAN & 59.65 & 83.48 & \textbf{81.55} \\
& KAN+Guidance & 54.64 & 84.18 & 81.55 \\
\midrule

\multirow{4}{*}{0.75}
& KAN & 58.07 & 83.87 & 81.54 \\
& KAN+Guidance & 54.31 & 84.47 & 81.54 \\
& TA-KAN & 55.78 & 84.61 & 81.47 \\
& TA-KAN+Guidance & 51.46 & 85.28 & 81.47 \\
\midrule

\multirow{4}{*}{0.5}
& KAN & 57.85 & 84.23 & 80.78 \\
& KAN+Guidance & 54.47 & 84.76 & 80.78 \\
& TA-KAN & 52.11 & 85.86 & 80.86 \\
& TA-KAN+Guidance & \textbf{48.73} & \textbf{86.42} & {80.86} \\
\midrule

\multirow{4}{*}{0.25}
& KAN & 57.90 & 83.75 & 77.33 \\
& KAN+Guidance & 57.01 & 84.03 & 77.33 \\
& TA-KAN & 50.22 & 85.82 & 78.43 \\
& TA-KAN+Guidance & 49.13 & 86.15 & 78.43 \\
\bottomrule
\end{tabular}
}
\label{tab:alpha_fixed}
\vspace{-10pt}
\end{wraptable}
\textbf{Effect of fixed $\alpha$ in standard KAN.}
We compare the performance of the standard KAN with a fixed RBF width (controlled by $\alpha$) across different values, as shown in \cref{tab:alpha_fixed}. 
We observe that simply tuning $\alpha$ does not lead to consistent improvements in OOD detection or ID classification performance.
For a fair comparison, we also report the performance of TA-KAN using the same values as $\alpha_{\min}$. 
In contrast to the standard KAN, TA-KAN consistently improves performance across all settings.
These results indicate that globally adjusting the RBF width is insufficient; instead, task-adaptive control of locality is crucial. 
This highlights that the performance gains of TA-KAN stem from its ability to dynamically modulate locality across tasks, rather than from a particular choice of $\alpha$.
See \cref{app:more_ablation} for detailed results and additional ablation studies.

\section{Conclusion}
\vspace{-2.5pt}

We studied OOD detection in rehearsal-free continual learning, where both catastrophic forgetting and lack of past data make reliable OOD detection challenging. We showed that standard KAN classifiers suffer from recency bias, degrading OOD performance for older tasks. To address this, we proposed TA-KAN, a task-adaptive classifier that improves ID–OOD separability via adaptive locality, together with a geometry-guided OOD score that leverages prompt-space structure without requiring past data. Our method consistently improves OOD detection while preserving ID accuracy, achieving state-of-the-art performance with negligible overhead.

\newpage

\begin{ack}
This research used the LIEF HPC-GPGPU Facility at the University of Melbourne, established with LIEF Grant LE170100200.
Sarah Erfani is in part supported by Australian Research Council (ARC) Discovery Early Career Researcher Award (DECRA) DE220100680.
\end{ack}

\bibliographystyle{plainnat}
\begingroup
\small
\setlength{\itemsep}{-1pt}
\setlength{\labelsep}{0.3em}   
\setlength{\parskip}{0pt}
\setlength{\parsep}{0pt}
\bibliography{main}

@String(CVPR  = {IEEE Conf. Comput. Vis. Pattern Recog.})

@String(ICCV  = {Int. Conf. Comput. Vis.})

@String(ECCV  = {Eur. Conf. Comput. Vis.})

@String(NeurIPS = {Adv. Neural Inform. Process. Syst.})

@String(ICML  = {Int. Conf. Mach. Learn.})

@String(ICLR  = {Int. Conf. Learn. Represent.})

@String(CVPRW = {IEEE Conf. Comput. Vis. Pattern Recog. Worksh.})

@String(AAAI  = {AAAI})

@String(WACV = {IEEE Winter Conf. Appl. Comput. Vis.})

@String(ICDM = {IEEE International Conference on Data Mining})

@String(CVPR  = {CVPR})

@String(ICCV  = {ICCV})

@String(ECCV  = {ECCV})

@String(NeurIPS = {NeurIPS})

@String(ICML  = {ICML})

@String(ICLR  = {ICLR})

@String(CVPRW = {CVPRW})

@String(WACV  = {WACV})

@String(ICDM  = {ICDM})

@inproceedings{NNGuide,
  title={Nearest neighbor guidance for out-of-distribution detection},
  author={Park, Jaewoo and Jung, Yoon Gyo and Teoh, Andrew Beng Jin},
  booktitle=ICCV,
  pages={1686--1695},
  year={2023}
}

@article{healtheg,
  title={Potential, challenges and future directions for deep learning in prognostics and health management applications},
  author={Fink, Olga and Wang, Qin and Svensen, Markus and Dersin, Pierre and Lee, Wan-Jui and Ducoffe, Melanie},
  journal={Engineering Applications of Artificial Intelligence},
  volume={92},
  pages={103678},
  year={2020},
  publisher={Elsevier}
}

@inproceedings{hendrycks_2017MSP,
  author       = {Dan Hendrycks and
                  Kevin Gimpel},
  title        = {A Baseline for Detecting Misclassified and Out-of-Distribution Examples
                  in Neural Networks},
  booktitle    = ICLR,
  year         = {2017},
}

@inproceedings{EnergyOE,
  title={Energy-based out-of-distribution detection},
  author={Liu, Weitang and Wang, Xiaoyun and Owens, John and Li, Yixuan},
  booktitle=NeurIPS,
  pages={21464--21475},
  year={2020}
}

@inproceedings{MetaOOD,
  title={Entropy maximization and meta classification for out-of-distribution detection in semantic segmentation},
  author={Chan, Robin and Rottmann, Matthias and Gottschalk, Hanno},
  booktitle=ICCV,
  pages={5128--5137},
  year={2021}
}

@InProceedings{MaxLogit,
  title = 	 {Scaling Out-of-Distribution Detection for Real-World Settings},
  author =       {Hendrycks, Dan and Basart, Steven and Mazeika, Mantas and Zou, Andy and Kwon, Joseph and Mostajabi, Mohammadreza and Steinhardt, Jacob and Song, Dawn},
  booktitle = 	 ICML,
  pages = 	 {8759--8773},
  year = 	 {2022}
}

@inproceedings{KNN_OOD,
  title={Out-of-distribution detection with deep nearest neighbors},
  author={Sun, Yiyou and Ming, Yifei and Zhu, Xiaojin and Li, Yixuan},
  booktitle=ICML,
  pages={20827--20840},
  year={2022},
  organization={PMLR}
}

@inproceedings{lee_2018_mahalanobis,
  title={A simple unified framework for detecting out-of-distribution samples and adversarial attacks},
  author={Lee, Kimin and Lee, Kibok and Lee, Honglak and Shin, Jinwoo},
  booktitle=NeurIPS,
  year={2018},
}

@inproceedings{hendrycks_2018OE,
  author       = {Dan Hendrycks and
                  Mantas Mazeika and
                  Thomas G. Dietterich},
  title        = {Deep Anomaly Detection with Outlier Exposure},
  booktitle    = ICLR,
  year         = {2019},
}

@inproceedings{VIM,
  title={Vim: Out-of-distribution with virtual-logit matching},
  author={Wang, Haoqi and Li, Zhizhong and Feng, Litong and Zhang, Wayne},
  booktitle=CVPR,
  pages={4921--4930},
  year={2022}
}

@article{outliersynthesis,
  title={Vos: Learning what you don't know by virtual outlier synthesis},
  author={Du, Xuefeng and Wang, Zhaoning and Cai, Mu and Li, Yixuan},
  journal={arXiv preprint arXiv:2202.01197},
  year={2022}
}

@inproceedings{CIL1st,
  title={icarl: Incremental classifier and representation learning},
  author={Rebuffi, Sylvestre-Alvise and Kolesnikov, Alexander and Sperl, Georg and Lampert, Christoph H},
  booktitle=CVPR,
  pages={2001--2010},
  year={2017}
}

@article{CL_survey,
  title={A comprehensive study of class incremental learning algorithms for visual tasks},
  author={Belouadah, Eden and Popescu, Adrian and Kanellos, Ioannis},
  journal={Neural Networks},
  volume={135},
  pages={38--54},
  year={2021},
  publisher={Elsevier}
}

@article{CL_survey2,
  title={A continual learning survey: Defying forgetting in classification tasks},
  author={De Lange, Matthias and Aljundi, Rahaf and Masana, Marc and Parisot, Sarah and Jia, Xu and Leonardis, Ale{\v{s}} and Slabaugh, Gregory and Tuytelaars, Tinne},
  journal={IEEE transactions on pattern analysis and machine intelligence},
  volume={44},
  number={7},
  pages={3366--3385},
  year={2021},
  publisher={IEEE}
}

@inproceedings{coshead,
  title={Learning a unified classifier incrementally via rebalancing},
  author={Hou, Saihui and Pan, Xinyu and Loy, Chen Change and Wang, Zilei and Lin, Dahua},
  booktitle=CVPR,
  pages={831--839},
  year={2019}
}

@inproceedings{CODA_prompt,
  title={Coda-prompt: Continual decomposed attention-based prompting for rehearsal-free continual learning},
  author={Smith, James Seale and Karlinsky, Leonid and Gutta, Vyshnavi and Cascante-Bonilla, Paola and Kim, Donghyun and Arbelle, Assaf and Panda, Rameswar and Feris, Rogerio and Kira, Zsolt},
  booktitle=CVPR,
  pages={11909--11919},
  year={2023}
}

@inproceedings{Dual_prompt,
  title={Dualprompt: Complementary prompting for rehearsal-free continual learning},
  author={Wang, Zifeng and Zhang, Zizhao and Ebrahimi, Sayna and Sun, Ruoxi and Zhang, Han and Lee, Chen-Yu and Ren, Xiaoqi and Su, Guolong and Perot, Vincent and Dy, Jennifer and others},
  booktitle=ECCV,
  pages={631--648},
  year={2022},
  organization={Springer}
}

@inproceedings{L2P,
  title={Learning to prompt for continual learning},
  author={Wang, Zifeng and Zhang, Zizhao and Lee, Chen-Yu and Zhang, Han and Sun, Ruoxi and Ren, Xiaoqi and Su, Guolong and Perot, Vincent and Dy, Jennifer and Pfister, Tomas},
  booktitle=CVPR,
  pages={139--149},
  year={2022}
}

@article{ER,
  title={Experience replay for continual learning},
  author={Rolnick, David and Ahuja, Arun and Schwarz, Jonathan and Lillicrap, Timothy and Wayne, Gregory},
  journal=NeurIPS,
  volume={32},
  year={2019}
}

@article{GEM,
  title={Gradient episodic memory for continual learning},
  author={Lopez-Paz, David and Ranzato, Marc'Aurelio},
  journal=NeurIPS,
  volume={30},
  year={2017}
}

@inproceedings{KAC,
  title={Kac: Kolmogorov-arnold classifier for continual learning},
  author={Hu, Yusong and Liang, Zichen and Yang, Fei and Hou, Qibin and Liu, Xialei and Cheng, Ming-Ming},
  booktitle=CVPR,
  pages={15297--15307},
  year={2025}
}

@article{shannonEntropy,
  title={A mathematical theory of communication},
  author={Shannon, Claude E},
  journal={The Bell system technical journal},
  volume={27},
  number={3},
  pages={379--423},
  year={1948},
  publisher={Nokia Bell Labs}
}

@inproceedings{syntheticOOD,
  title={Mixture outlier exposure: Towards out-of-distribution detection in fine-grained environments},
  author={Zhang, Jingyang and Inkawhich, Nathan and Linderman, Randolph and Chen, Yiran and Li, Hai},
  booktitle=WACV,
  pages={5531--5540},
  year={2023}
}

@inproceedings{GramMatrix,
  title={Detecting out-of-distribution examples with gram matrices},
  author={Sastry, Chandramouli Shama and Oore, Sageev},
  booktitle=ICML,
  pages={8491--8501},
  year={2020},
  organization={PMLR}
}

@article{retrainOOD,
  title={Towards maximizing the representation gap between in-domain \& out-of-distribution examples},
  author={Nandy, Jay and Hsu, Wynne and Lee, Mong Li},
  journal=NeurIPS,
  volume={33},
  pages={9239--9250},
  year={2020}
}

@inproceedings{retrainOOD2,
  title={Energy-based open-world uncertainty modeling for confidence calibration},
  author={Wang, Yezhen and Li, Bo and Che, Tong and Zhou, Kaiyang and Liu, Ziwei and Li, Dongsheng},
  booktitle=ICCV,
  pages={9302--9311},
  year={2021}
}

@inproceedings{KAN,
  title={Kan: Kolmogorov-arnold networks},
  author={Liu, Ziming and Wang, Yixuan and Vaidya, Sachin and Ruehle, Fabian and Halverson, James and Solja{\v{c}}i{\'c}, Marin and Hou, Thomas Y and Tegmark, Max},
  booktitle=ICLR,
  year={2025}
}

@article{RBFKAN,
  title={Kolmogorov-arnold networks are radial basis function networks},
  author={Li, Ziyao},
  journal={arXiv preprint arXiv:2405.06721},
  year={2024}
}

@article{RBF,
  title={Radial basis functions for multivariable interpolation: a review},
  author={Powell, Michael JD},
  journal={Algorithms for approximation},
  pages={143--167},
  year={1987}
}

@article{CUB,
  title={The caltech-ucsd birds-200-2011 dataset},
  author={Wah, Catherine and Branson, Steve and Welinder, Peter and Perona, Pietro and Belongie, Serge},
  year={2011}
}

@inproceedings{OODCILunsuper,
  title={Out-of-distribution detection in unsupervised continual learning},
  author={He, Jiangpeng and Zhu, Fengqing},
  booktitle=CVPRW,
  pages={3850--3855},
  year={2022}
}

@article{MORE1,
  title={Open-world continual learning: Unifying novelty detection and continual learning},
  author={Kim, Gyuhak and Xiao, Changnan and Konishi, Tatsuya and Ke, Zixuan and Liu, Bing},
  journal={Artificial Intelligence},
  volume={338},
  pages={104237},
  year={2025},
  publisher={Elsevier}
}

@book{KAT,
  title={On the representation of continuous functions of several variables by superpositions of continuous functions of a smaller number of variables},
  author={Kolmogorov, Andrei Nikolaevich},
  year={1961},
  publisher={American Mathematical Society}
}

@inproceedings{Bspline,
  title={General matrix representations for B-splines},
  author={Qin, Kaihuai},
  booktitle={Proceedings Pacific Graphics' 98. Sixth Pacific Conference on Computer Graphics and Applications (Cat. No. 98EX208)},
  pages={37--43},
  year={1998},
  organization={IEEE}
}

@article{RBFKAN2,
  title={RBF-KAN: Radial Basis Function-Kolmogorov-Arnold Networks},
  author={Chao, Zengfu and Liu, Xinran and Wu, Zheyu and Li, Xiaoping},
  journal={IEEE Internet of Things Journal},
  year={2026},
  publisher={IEEE}
}

@inproceedings{H2ST,
  title={H2ST: Hierarchical two-sample tests for continual out-of-distribution detection},
  author={Liu, Yuhang and Zhao, Wenjie and Guo, Yunhui},
  booktitle=CVPR,
  pages={15413--15423},
  year={2025}
}

@inproceedings{MORE,
  title={A multi-head model for continual learning via out-of-distribution replay},
  author={Kim, Gyuhak and Liu, Bing and Ke, Zixuan},
  booktitle={Conference on lifelong learning agents},
  pages={548--563},
  year={2022},
  organization={PMLR}
}

@inproceedings{CPrompt,
  title={Consistent prompting for rehearsal-free continual learning},
  author={Gao, Zhanxin and Cen, Jun and Chang, Xiaobin},
  booktitle=CVPR,
  pages={28463--28473},
  year={2024}
}

@inproceedings{CIL_KD,
  title={Class incremental learning with multi-teacher distillation},
  author={Wen, Haitao and Pan, Lili and Dai, Yu and Qiu, Heqian and Wang, Lanxiao and Wu, Qingbo and Li, Hongliang},
  booktitle=CVPR,
  pages={28443--28452},
  year={2024}
}

@article{vit,
  title={An image is worth 16x16 words: Transformers for image recognition at scale},
  author={Dosovitskiy, Alexey and Beyer, Lucas and Kolesnikov, Alexander and Weissenborn, Dirk and Zhai, Xiaohua and Unterthiner, Thomas and Dehghani, Mostafa and Minderer, Matthias and Heigold, Georg and Gelly, Sylvain and others},
  journal={arXiv preprint arXiv:2010.11929},
  year={2020}
}

@inproceedings{parsimony_CIL,
  title={Learn-prune-share for lifelong learning},
  author={Wang, Zifeng and Jian, Tong and Chowdhury, Kaushik and Wang, Yanzhi and Dy, Jennifer and Ioannidis, Stratis},
  booktitle=ICDM,
  pages={641--650},
  year={2020},
  organization={IEEE}
}

@article{Bisecle,
  title={Bisecle: Binding and separation in continual learning for video language understanding},
  author={Tan, Yue and Hu, Xiaoqian and Xue, Hao and De Melo, Celso and Salim, Flora D},
  journal={arXiv preprint arXiv:2507.00469},
  year={2025}
}

@inproceedings{SupLID,
  title={SupLID: Geometrical Guidance for Out-of-Distribution Detection in Semantic Segmentation},
  author={Udayangani, Nimeshika and Erfani, Sarah and Leckie, Christopher},
  booktitle={Proceedings of the 34th ACM International Conference on Information and Knowledge Management},
  pages={2905--2914},
  year={2025}
}

@inproceedings{ImageNet-R,
  title={The many faces of robustness: A critical analysis of out-of-distribution generalization},
  author={Hendrycks, Dan and Basart, Steven and Mu, Norman and Kadavath, Saurav and Wang, Frank and Dorundo, Evan and Desai, Rahul and Zhu, Tyler and Parajuli, Samyak and Guo, Mike and others},
  booktitle=ICCV,
  pages={8340--8349},
  year={2021}
}

@article{cifar,
  title={Learning multiple layers of features from tiny images},
  author={Krizhevsky, Alex and Hinton, Geoffrey and others},
  year={2009},
  publisher={Toronto, ON, Canada}
}

@inproceedings{proKT,
  title={Learning to prompt knowledge transfer for open-world continual learning},
  author={Li, Yujie and Yang, Xin and Wang, Hao and Wang, Xiangkun and Li, Tianrui},
  booktitle=AAAI,
  volume={38},
  number={12},
  pages={13700--13708},
  year={2024}
}

@incollection{catasforgetting,
  title={Catastrophic interference in connectionist networks: The sequential learning problem},
  author={McCloskey, Michael and Cohen, Neal J},
  booktitle={Psychology of learning and motivation},
  volume={24},
  pages={109--165},
  year={1989},
  publisher={Elsevier}
}

@inproceedings{syntheticreplayCIL,
  title={Ddgr: Continual learning with deep diffusion-based generative replay},
  author={Gao, Rui and Liu, Weiwei},
  booktitle=ICML,
  pages={10744--10763},
  year={2023},
  organization={PMLR}
}

@inproceedings{syntheticreplayCIL2,
  title={Sddgr: Stable diffusion-based deep generative replay for class incremental object detection},
  author={Kim, Junsu and Cho, Hoseong and Kim, Jihyeon and Tiruneh, Yihalem Yimolal and Baek, Seungryul},
  booktitle=CVPR,
  pages={28772--28781},
  year={2024}
}

@article{syntheticreplayCIL3,
  title={Continual learning with deep generative replay},
  author={Shin, Hanul and Lee, Jung Kwon and Kim, Jaehong and Kim, Jiwon},
  journal=NeurIPS,
  volume={30},
  year={2017}
}

@article{LLMs,
  title={Llama: Open and efficient foundation language models},
  author={Touvron, Hugo and Lavril, Thibaut and Izacard, Gautier and Martinet, Xavier and Lachaux, Marie-Anne and Lacroix, Timoth{\'e}e and Rozi{\`e}re, Baptiste and Goyal, Naman and Hambro, Eric and Azhar, Faisal and others},
  journal={arXiv preprint arXiv:2302.13971},
  year={2023}
}

@article{S_Prompts,
  title={S-prompts learning with pre-trained transformers: An occam’s razor for domain incremental learning},
  author={Wang, Yabin and Huang, Zhiwu and Hong, Xiaopeng},
  journal=NeurIPS,
  volume={35},
  pages={5682--5695},
  year={2022}
}

@article{HidePromt,
  title={Hierarchical decomposition of prompt-based continual learning: Rethinking obscured sub-optimality},
  author={Wang, Liyuan and Xie, Jingyi and Zhang, Xingxing and Huang, Mingyi and Su, Hang and Zhu, Jun},
  journal=NeurIPS,
  volume={36},
  pages={69054--69076},
  year={2023}
}

@article{AIauto,
  title={AI autonomy: Self-initiated open-world continual learning and adaptation},
  author={Liu, Bing and Mazumder, Sahisnu and Robertson, Eric and Grigsby, Scott},
  journal={AI Magazine},
  volume={44},
  number={2},
  pages={185--199},
  year={2023},
  publisher={Wiley Online Library}
}

@inproceedings{ORE,
  title={Towards open world object detection},
  author={Joseph, KJ and Khan, Salman and Khan, Fahad Shahbaz and Balasubramanian, Vineeth N},
  booktitle=CVPR,
  pages={5830--5840},
  year={2021}
}

@article{TIL,
  title={A continual learning survey: Defying forgetting in classification tasks},
  author={De Lange, Matthias and Aljundi, Rahaf and Masana, Marc and Parisot, Sarah and Jia, Xu and Leonardis, Ale{\v{s}} and Slabaugh, Gregory and Tuytelaars, Tinne},
  journal={IEEE transactions on pattern analysis and machine intelligence},
  volume={44},
  number={7},
  pages={3366--3385},
  year={2021},
  publisher={IEEE}
}

@article{c2st,
  title={Revisiting classifier two-sample tests},
  author={Lopez-Paz, David and Oquab, Maxime},
  journal={arXiv preprint arXiv:1610.06545},
  year={2016}
}

@article{MLP,
  title={Multilayer feedforward networks are universal approximators},
  author={Hornik, Kurt and Stinchcombe, Maxwell and White, Halbert},
  journal={Neural networks},
  volume={2},
  number={5},
  pages={359--366},
  year={1989},
  publisher={Elsevier}
}

@article{fkan,
  title={fkan: Fractional kolmogorov--arnold networks with trainable jacobi basis functions},
  author={Aghaei, Alireza Afzal},
  journal={Neurocomputing},
  volume={623},
  pages={129414},
  year={2025},
  publisher={Elsevier}
}

@article{KAN_survey,
  title={A survey on kolmogorov-arnold network},
  author={Somvanshi, Shriyank and Javed, Syed Aaqib and Islam, Md Monzurul and Pandit, Diwas and Das, Subasish},
  journal={ACM Computing Surveys},
  volume={58},
  number={2},
  pages={1--35},
  year={2025},
  publisher={ACM New York, NY}
}

@inproceedings{KAN_OOD,
title={Advancing Out-of-Distribution Detection via Local Neuroplasticity},
author={Alessandro Canevaro and Julian Schmidt and Mohammad Sajad Marvi and Hang Yu and Georg Martius and Julian Jordan},
booktitle=ICLR,
year={2025},
url={https://openreview.net/forum?id=1F8xTfv6ah}
}

@inproceedings{UCL,
  title={Unsupervised continual learning via pseudo labels},
  author={He, Jiangpeng and Zhu, Fengqing},
  booktitle={International Workshop on Continual Semi-Supervised Learning},
  pages={15--32},
  year={2021},
  organization={Springer}
}

@article{freeKAN,
  title={Free-RBF-KAN: Kolmogorov-Arnold Networks with Adaptive Radial Basis Functions for Efficient Function Learning},
  author={Chiu, Shao-Ting and Cheung, Siu Wun and Braga-Neto, Ulisses and Lee, Chak Shing and Li, Rui Peng},
  journal={arXiv preprint arXiv:2601.07760},
  year={2026}
}
\endgroup

\newpage
\appendix

\begin{figure}[tbh!]
\vspace{-15pt}
    \centering
    \begin{subfigure}[t]{0.25\linewidth}
        \centering
        \includegraphics[width=\linewidth]{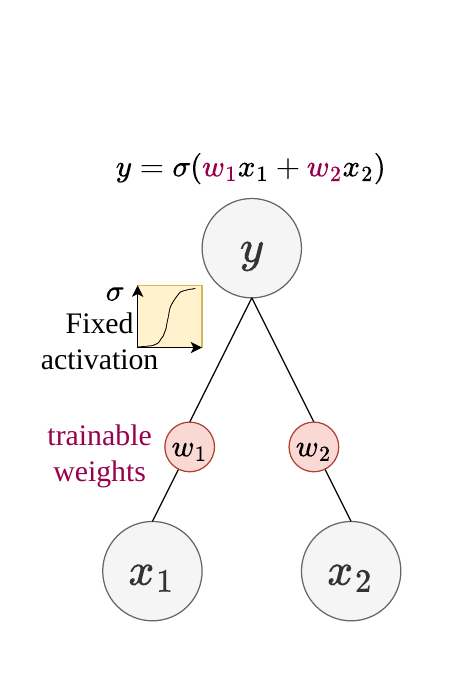}
        \vspace{-25pt}
        \caption{Multi Layer Perceptron (MLP)}
        \label{fig:first}
    \end{subfigure}
    \hspace{0.08\linewidth}
    \begin{subfigure}[t]{0.27\linewidth}
        \centering
        \includegraphics[width=\linewidth]{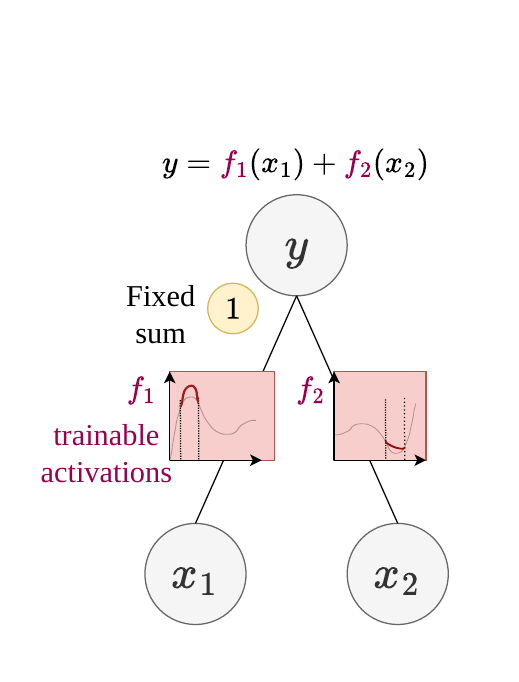}
        \vspace{-25pt}
        \caption{Kolmogorov-Arnold Networks (KAN)}
        \label{fig:second}
    \end{subfigure}
\caption{Simplified illustration of KAN~\cite{KAN} and MLP~\cite{MLP} architectures.
(a) MLP uses globally shared weights with fixed activation functions, where all parameters are equally updated across all tasks, resulting in updates on irrelevant regions during new task learning. 
(b) In contrast, KAN employs learnable activation functions, enabling localized responses where only input-relevant regions are activated and updated, thereby preventing impact on other regions and mitigating catastrophic forgetting.}
\vspace{-15pt}
    \label{fig:kan_vs_mlp}
\end{figure}

\section{Reproducibility} \label{app:reproducibility}
The experimental settings and implementation details of our method are reported in Sec.~\ref{sec:ex_settings} and \cref{app:configuration}. 
Our code is publicly available at \url{https://github.com/hdnugit/TAKAN/}. 
All experiments are conducted on an NVIDIA A100 GPU. 
Detailed information on the hardware and software environment is also provided in the repository.

\section{Related Work}
\label{app:related_work}

\textbf{Class-Incremental Learning (CIL)} aims to learn a sequence of classification tasks, requiring the model to perform prediction over all previously encountered classes. 
However, catastrophic forgetting~\cite{catasforgetting} remains a fundamental challenge in CIL. 
To address this, prior methods typically employ \textit{rehearsal-based} strategies, which retain a small subset of past data~\cite{ER,GEM} or generate synthetic samples~\cite{syntheticreplayCIL,syntheticreplayCIL2,syntheticreplayCIL3}, and combine them with new-task data to train the model.
However, such approaches raise practical limitations, as long-term storage of training data will violate data privacy and incur high memory costs~\cite{L2P}.
In contrast, recent advancements in pre-trained models and large language models (LLMs) \cite{LLMs} have enabled prompt-based methods, which have emerged as state-of-the-art ~\cite{CODA_prompt,CPrompt,Bisecle,L2P,Dual_prompt} and provide a rehearsal-free alternative for CIL.
Instead of directly modifying encoder weights on newly arriving data, these methods introduce a small set of learnable parameters (prompts) to instruct the model on sequentially arriving tasks, thereby reducing forgetting and encoding task-specific knowledge in a more succinct way.
Originally introduced in Natural Language Processing (NLP), prompt learning was first adapted to vision-based CIL by L2P~\cite{L2P}, while DualPrompt~\cite{Dual_prompt} further demonstrates that such methods can outperform advanced continual learning approaches that rely on relatively large rehearsal buffers.
Subsequent works, including S-Prompts~\cite{S_Prompts}, CODA-Prompt~\cite{CODA_prompt}, HiDe-Prompt~\cite{HidePromt}, Bisecle~\cite{Bisecle} and CPrompt~\cite{CPrompt}, explore diverse prompt learning strategies to improve knowledge retention and robustness in CIL.

Nevertheless, almost all existing CIL approaches~\cite{ER, CODA_prompt, Dual_prompt, L2P} operate under the closed world assumption, wherein test samples are assumed to be drawn independently and identically distributed~(i.i.d.) from the training distribution observed so far.
However, it is inevitable that continual learning systems encounter novel, unknown samples (OOD)—when deployed in open-world settings.
As a crucial first step, the model should reliably detect such samples rather than misclassifying them as one of the known classes, especially in safety-critical deployments such as healthcare~\cite{healtheg}.
Then, in a future task, when the model approaches their ground truth, these samples should be incrementally incorporated as in-distribution without forgetting previously learned knowledge~\cite{MORE1,proKT}.

\textbf{Out-of-Distribution~(OOD) Detection.}
A standard approach to OOD detection is to leverage confidence scores derived from deep neural classifiers, such as the maximum softmax probability (MSP)~\cite{hendrycks_2017MSP}, Energy~\cite{EnergyOE}—which also mirrors the class-conditional probability—Entropy (Entropy)~\cite{MetaOOD}, and the maximum of logit (MaxLogit)~\cite{MaxLogit}. 
However, classifier-based scores may suffer from overconfidence issues: while they assign low scores to samples near decision boundaries (useful for near-OOD), they can assign high confidence to far-OOD samples~\cite{NNGuide}. 
In contrast, methods such as kNN~\cite{KNN_OOD}, Mahalanobis~\cite{lee_2018_mahalanobis}, and LID~\cite{SupLID} rely on \textit{geometry-based} metrics in feature space for OOD detection. 
Since these approaches are based on the distance to ID data or local density in the feature space, they can reliably assign low scores to far-OOD regions~\cite{NNGuide}.
NNGuide~\cite{NNGuide} integrates both classifier-based and geometry-based signals to improve the robustness of OOD detection.
In contrast to the above post-hoc methods, \textit{retraining-based} approaches~\cite{hendrycks_2018OE,outliersynthesis} aim to enhance representations for better ID–OOD separability.

However, all these methods are primarily developed and evaluated in static settings.
Directly incorporating these methods into the CL setting is fundamentally challenging and insufficient due to several limitations.
On the one hand, these methods do not address the challenges arising from catastrophic forgetting.
Maintaining reliable decision boundaries for both inter-task class separation and OOD detection becomes difficult as the notion of the distribution evolves over time; what was previously OOD may become ID as new tasks are learned. 
On the other hand, the absence of training data from previous tasks makes a large class of methods—including geometry-based and retraining-based approaches—inapplicable, as they rely on access to all training data or require modifications to the training procedure.

\textbf{Out-of-distribution detection in Class-Incremental Learning (OOD-CL)} is a synergistic integration of continual learning and out-of-distribution detection~\cite{proKT}. 
It aims to deploy CIL models to operate in an open-world setting where the model must (i) identify samples that do not belong to the currently learned classes and (ii) incrementally incorporate such novel data into its knowledge base, while mitigating catastrophic forgetting.
In practice, incorporating detected unknowns requires access to labeled training data, which can be obtained through interaction with the environment or human users~\cite{MORE1}.
Following prior work~\cite{H2ST,proKT,MORE1,ORE,MORE}, we do not consider this data acquisition process~\cite{AIauto}, as it lies outside the scope of this study. Instead, we focus on the core learning challenges of OOD detection and class-incremental learning.

Although there exists a large body of work addressing OOD detection and CIL separately, only a few studies attempt to address OOD-CL.
In particular, \cite{MORE1} theoretically demonstrates the necessity of OOD detection in CIL.
ORE~\cite{ORE} proposes a contrastive clustering to enhance the separability between ID and OOD samples. 
However, it maintains a feature buffer for each past class to compute class prototypes at every iteration, which violates the parsimony~\cite{parsimony_CIL} in CIL~\cite{parsimony_CIL}.
Moreover, it relies on a standard energy-based score~\cite{EnergyOE} for OOD identification and adopts a rehearsal-based CIL framework to mitigate catastrophic forgetting.
Similarly, ProKT~\cite{proKT} applies the MaxLogit~\cite{MaxLogit} score to a generic prompt-based CIL model, without explicitly addressing the challenges of OOD-CL, instead combining an existing post-hoc score with a generic prompt-based CIL framework that maintains $M$ task-specific prompts.
Different from the standard CIL setting, MORE~\cite{MORE} employs a multi-head classifier and trains each task-specific model to distinguish between ID and OOD samples. 
However, it relies on stored samples in a rehearsal buffer as explicit OOD data to improve current task learning.
Different from these approaches, H2ST~\cite{H2ST} addresses OOD detection in the task-incremental learning (TIL) setting~\cite{TIL}, but similarly maintains a separate model for each task and relies on a rehearsal buffer to train C2ST~\cite{c2st} classifiers for each task head.
Moreover, MSP-BC-CE~\cite{OODCILunsuper} formulates OOD detection in unsupervised CIL (UCL)~\cite{UCL} as a post-hoc confidence calibration problem, and relies solely on classifier outputs whereas we address it through model-level design and geometric guidance, enabling more robust OOD detection in CIL.

Overall, existing approaches either naively couple standard OOD methods with CIL models~\cite{MORE1,proKT,ORE} without addressing the core challenges of applying OOD methods in the continual setting, or depend on rehearsal buffers containing past data~\cite{MORE,H2ST,ORE}. 
In particular, as rehearsal-free CIL emerges as the de facto standard due to its scalability and privacy advantages, enabling reliable OOD detection in this setting is crucial for real-world deployment.
Therefore, we develop a novel method for OOD detection in rehearsal-free continual learning and propose a new OOD-CL scoring function.

\textbf{Kolmogorov–Arnold Networks (KAN)}~\cite{KAN} are a novel neural architecture based on the Kolmogorov–Arnold representation theorem (KAT)~\cite{KAT}. 
Unlike traditional multi-layer perceptrons (MLPs)~\cite{MLP}, KANs incorporate learnable activation functions instead of fixed activations with shared weights, thereby exhibiting a distinctive property of local neuroplasticity~\cite{KAN_OOD}. 
The original KAN~\cite{KAN} employs nonlinear B-spline basis functions~\cite{Bspline} for function approximation, while subsequent studies~\cite{fkan,RBFKAN,RBFKAN2} explore alternative basis functions to improve performance.
RBF-KAN~\cite{RBFKAN} leverages univariate radial basis functions (RBFs)~\cite{RBF} within the superpositional structure of the KAT to approximate multivariate functions. 
They enhance the geometric interpretability and scalability of KAN, and more effectively address the curse of dimensionality in high-dimensional settings~\cite{RBFKAN2,KAC}.
While many works explore KAN across various domains~\cite{KAN_survey}, KAC~\cite{KAC} is the first to study KAN for mitigating catastrophic forgetting in CIL frameworks, demonstrating improved closed-world CIL performance over MLP-based networks. 
Motivated by this, we extend the study of KAN to the open-world CIL setting and propose a framework that can be integrated into generic rehearsal-free prompt-based models to improve their robustness.
It is worth noting that~\cite{KAN_OOD} leverages KAN as a post-hoc OOD detector, whereas our usage is fundamentally different. 
Specifically, their method operates independently of the backbone by comparing activations of a trained and an untrained KAN to derive an OOD score. 
Moreover, it relies on multiple KAN models (one per class) and aggregates their scores via a max operation. 
In contrast, we study KAN within the continual learning setting as an integral network component (i.e., the classifier head), enabling end-to-end learning without maintaining multiple detectors. 
This distinction is critical, as the post-hoc multi-model design is computationally expensive and does not scale to CIL, where the number of tasks or classes grows over time.

\section{Baselines and Configuration} \label{app:configuration}
We evaluate our method using five OOD detection methods: MSP~\cite{hendrycks_2017MSP}, MaxLogit~\cite{MaxLogit}, Energy~\cite{EnergyOE}, Entropy~\cite{MetaOOD}, KL~\cite{MaxLogit} and three OOD-CL methods: MORE~\cite{MORE}, MSP-BC-CE~\cite{OODCILunsuper} and Pro-KT~\cite{proKT}.
For MORE, which employs multi-head classifiers, we replace the linear heads with KAN-based heads to construct a stronger baseline.
Since MORE uses a rehearsal buffer as OOD training data for the current task, we implement it with a buffer size of 1000 across all datasets.
Similarly, for ProKT, which is also prompt-based, we replace its linear classifier with a KAN-based classifier of the same size as ours.
For MSP-BC-CE, originally proposed for OOD detection in unsupervised continual learning (UCL)~\cite{UCL}, we implement it on the same prompt-based model with the KAN head to establish a strong and fair baseline.
This provides a strong bias-correction baseline by explicitly compensating for class-weight norm imbalance in the KAN classifier.
We additionally consider two canonical classifier alternatives for mitigating classifier bias: a cosine-normalized classifier (\textit{Cosine-Head})~\cite{coshead} and a nearest-class-mean classifier (\textit{NCM Readout})~\cite{CIL1st}, both implemented within the same prompt-learning framework. 
For Cosine-Head, we replace the linear classifier with a cosine-normalized classifier, where the feature and class-weight vectors are $\ell_2$-normalized and classification is performed based on their cosine similarity~\cite{coshead}. 
For NCM Readout, prompts are trained using the same cosine-normalized classification objective, while inference is performed using cosine similarity to stored class prototypes~\cite{CIL1st}.
All reported results are averaged over three independent trial runs for each setting.
Following prior CL work~\cite{CODA_prompt,KAC}, each trial uses a different random class-order shuffle generated with a trial-specific seed (set to the loop index), from which the task sequence is constructed.
Within each trial, the class order and all other sources of randomness, including model initialization, are kept identical across methods to ensure a fair comparison.
Furthermore, we follow the original hyperparameter settings for each method where applicable, and keep all remaining hyperparameters consistent to ensure a fair comparison.
For TA-KAN, the hyperparameter $\alpha_{\min}$ is selected as 0.5 from the set $\{0.25, 0.5, 0.75\}$ using a small held-out validation set, following~\cite{ORE}, for all task settings except the 40-task setting, where we set $\alpha_{\min} = 0.25$.
We set $\alpha_{\max} = 1$ (default step-size width) to allow sufficient generalization for newer tasks.
All experiments are conducted using a ViT-B/16 backbone, following prior work~\cite{CODA_prompt,Dual_prompt,KAC}.

\section{More Experimental Results and Analysis} \label{app:more_results}


\textbf{Comparison with bias-mitigation classifier heads.}
We further analyze the Cosine-Head and NCM Readout baselines in \cref{tab:cub200_mainresults,tab:imagenetr_mainresults}, which provide alternative mechanisms for mitigating classifier bias in CIL.
While the Cosine Head consistently improves over the linear head, particularly in ID classification and often in OOD detection, both Cosine-Head and NCM Readout remain overall inferior to the KAN head on OOD detection, particularly on the challenging CUB-200 benchmark. 
We also compare with MSP-BC-CE, which addresses classifier bias in OOD-CL and is implemented on top of the KAN-Head in our experiments.
Although MSP-BC-CE improves OOD detection over the standard KAN-Head in several settings, TA-KAN achieves stronger overall OOD detection, with particularly substantial gains on fine-grained CUB-200.
Importantly, these results indicate that neither replacing the linear classifier with normalized or prototype-based decision rules nor applying bias correction to the standard KAN classifier is sufficient to explain the improvements achieved by TA-KAN.
Instead, TA-KAN modifies the classifier function itself through task-adaptive control of RBF locality, preserving KAN's local plasticity while improving ID--OOD separability for older tasks affected by recency bias.

\begin{table}[tbh!]
\centering
\caption{Results on CIFAR-100~\cite{cifar} dataset under different task settings.Lower FPR is better; higher AUC and ACC are better.}
\small
\renewcommand{\arraystretch}{1.1}
\setlength{\tabcolsep}{4pt}
\resizebox{0.9\textwidth}{!}{
\begin{tabular}{lccc|ccc|ccc}
\toprule
\multirow{2}{*}{\textbf{Method}} 
& \multicolumn{3}{c}{\textbf{5 Tasks}} 
& \multicolumn{3}{c}{\textbf{10 Tasks}} 
& \multicolumn{3}{c}{\textbf{20 Tasks}} \\
\cmidrule(lr){2-4} \cmidrule(lr){5-7} \cmidrule(lr){8-10}
& $FPR_L \downarrow$ & $AUC_L \uparrow$ & $ACC_L \uparrow$ 
& $FPR_L \downarrow$ & $AUC_L \uparrow$ & $ACC_L \uparrow$ 
& $FPR_L \downarrow$ & $AUC_L \uparrow$ & $ACC_L \uparrow$ \\
\midrule
Linear-Head~\cite{CODA_prompt}            & 36.45 & 90.67 & \textbf{88.71} & 41.58 & 89.30 & 86.55 & 49.92 & 86.27 & 82.96 \\
KAN-Head (Baseline)~\cite{KAC} & 37.28 & 89.73 & {87.92} & 42.79 & 88.34 & \textbf{86.71} & 45.06 & 86.84 & \textbf{84.83} \\
ProKT~\cite{proKT}               & 45.91 & 86.44 & 83.72 & 52.97 & 83.62 & 79.27 & 64.20 & 77.43 & 72.54 \\
MORE~\cite{MORE}                   & 36.95 & 89.17 & 85.67 & 48.10 & 86.94 & 85.43 & 51.32 & 79.24 & 74.27 \\
MSP-BC-CE~\cite{OODCILunsuper}              & \textbf{33.96} & \textbf{92.22} & 87.92 & 41.77 & 89.98 & 86.71 & 46.53 & 87.34 & 84.83 \\
\midrule
TA-KAN                & 35.96 & 90.85 & 87.63 & 40.92 & 89.55 & 86.53 & 45.68 & 87.12 & 84.04 \\
TA-KAN+Guidance
                      & {34.50} & {91.32} & 87.63 
                      & \textbf{39.66} & \textbf{90.02} & 86.53 
                      & \textbf{43.94} & \textbf{87.87} & 84.04 \\
\bottomrule
\end{tabular}
}
\label{tab:cifar100_main}
\end{table}
\textbf{Results on CIFAR-100.} 
We report results on CIFAR-100 dataset in \cref{tab:cifar100_main} across all task settings. 
Consistent with other datasets, replacing the standard KAN head with the proposed TA-KAN classifier improves OOD detection performance in terms of both FPR and AUC, while preserving the ID performance gains of KAN. 
Incorporating the proposed guidance into the TA-KAN scoring function further yields additional improvements in OOD performance. 
Note that the improvement in ID performance of the standard KAN over the linear head is relatively small compared to other datasets, and a similar trend is observed for TA-KAN. 
This trend is consistent with prior observations in KAN-based classifiers~\cite{KAC}, where performance gains tend to saturate on low-resolution datasets such as CIFAR-100 when using pre-trained backbones.

\begin{table}[tbh!]
\centering
\caption{Results on ImageNet-R~\cite{ImageNet-R} under different task settings (mean ± standard deviation over random trials). Lower FPR is better; higher AUC and ACC are better.}
\small
\renewcommand{\arraystretch}{1.0}
\setlength{\tabcolsep}{2pt}
\resizebox{0.99\textwidth}{!}{
\begin{tabular}{lccc|ccc|ccc}
\toprule
\multirow{2}{*}{\textbf{Method}}
& \multicolumn{3}{c}{\textbf{5 Tasks}}
& \multicolumn{3}{c}{\textbf{10 Tasks}}
& \multicolumn{3}{c}{\textbf{20 Tasks}} \\
\cmidrule(lr){2-4} \cmidrule(lr){5-7} \cmidrule(lr){8-10} 
& $FPR_L$ & $AUC_L$ & $ACC_L$
& $FPR_L$ & $AUC_L$ & $ACC_L$
& $FPR_L$ & $AUC_L$ & $ACC_L$ \\
\midrule

Linear-Head~\cite{CODA_prompt}
& $42.74 \pm 1.29$ & $85.44 \pm 0.39$ & $77.96 \pm 0.26$
& $45.69 \pm 2.71$ & $86.26 \pm 0.71$ & $76.51 \pm 0.32$
& $53.49 \pm 6.59$ & $85.28 \pm 1.63$ & $73.90 \pm 0.32$ \\

KAN-Head (Baseline)~\cite{KAC}
& $42.52 \pm 1.28$ & $84.81 \pm 0.32$ & $\textbf{79.25} \pm 0.38$
& $43.04 \pm 3.06$ & $85.64 \pm 0.28$ & $\textbf{78.76} \pm 0.26$
& $44.00 \pm 7.79$ & $85.40 \pm 1.17$ & $\textbf{77.42} \pm 0.30$ \\

MSP-BC-CE~\cite{OODCILunsuper}
& $40.65 \pm 1.52$ & $85.50 \pm 0.21$ & $79.25 \pm 0.38$
& $42.12 \pm 2.91$ & $\underline{86.84} \pm 0.74$ & $78.76 \pm 0.26$
& $44.88 \pm 7.04$ & $85.78 \pm 1.21$ & $77.42 \pm 0.30$ \\

\midrule

TA-KAN
& $\underline{40.07} \pm 1.18$ & $\underline{85.77} \pm 0.29$ & $78.83 \pm 0.22$
& $\underline{40.25} \pm 3.00$ & $86.76 \pm 0.48$ & $77.76 \pm 0.49$
& $\underline{42.57} \pm 8.16$ & $\underline{85.99} \pm 1.38$ & $77.03 \pm 0.10$ \\

TA-KAN + Guidance
& $\textbf{39.72} \pm 1.35$ & $\textbf{85.92} \pm 0.35$ & $\underline{78.83} \pm 0.22$
& $\textbf{39.98} \pm 2.90$ & $\textbf{86.87} \pm 0.48$ & $\underline{77.76} \pm 0.49$
& $\textbf{42.61} \pm 8.39$ & $\textbf{86.10} \pm 1.55$ & $\underline{77.03} \pm 0.10$ \\

\bottomrule
\end{tabular}
}
\label{tab:std_results_imagenet}
\end{table}
\textbf{Analysis of variance across trials.}
For each experimental setting, the main results report the mean over multiple independent trial runs.
We additionally report the corresponding standard deviations for ImageNet-R in \cref{tab:std_results_imagenet} to assess performance stability across trials.
Each trial uses a different random class order and task partition, in addition to optimization randomness.
Consequently, the reported standard deviations reflect robustness to both optimization variability and changes in the continual learning setting itself.
The consistently low standard deviations in AUROC and classification accuracy indicate stable OOD detection and continual learning performance across trials.
A modest increase in variance is observed as the number of tasks increases, which is expected as the continual learning setting becomes more challenging~\cite{CODA_prompt}.
FPR exhibits comparatively higher variability due to its greater sensitivity to the operating threshold, while this trend is observed across methods rather than being specific to TA-KAN.

\section{Ablation Study and Analysis} \label{app:more_ablation}

\subsection{Detailed Ablation Study Results of Different Components} \label{app:ablation_component}
\begin{table*}[tbh!]
\centering
\caption{Detailed ablation study results of different components. Results are reported over all task settings on CUB-200 dataset. Lower FPR is better; higher AUC and ACC are better.}
\small
\setlength{\tabcolsep}{0.0001pt}
\renewcommand{\arraystretch}{1.2}
\makebox[\textwidth][c]{
\resizebox{1.1\textwidth}{!}{
\begin{tabular}{lcccccc|cccccc|cccccc|cccccc}
\toprule
\multirow{2}{*}{\textbf{Method}} 
& \multicolumn{6}{c}{\textbf{5 Tasks}} 
& \multicolumn{6}{c}{\textbf{10 Tasks}} 
& \multicolumn{6}{c}{\textbf{20 Tasks}} 
& \multicolumn{6}{c}{\textbf{Average}} \\
\cmidrule(lr){2-7} \cmidrule(lr){8-13} \cmidrule(lr){14-19} \cmidrule(lr){20-25}
& $FPR_L$ & $FPR_A$ & $AUC_L$ & $AUC_A$ & $ACC_L$ & $ACC_A$
& $FPR_L$ & $FPR_A$ & $AUC_L$ & $AUC_A$ & $ACC_L$ & $ACC_A$
& $FPR_L$ & $FPR_A$ & $AUC_L$ & $AUC_A$ & $ACC_L$ & $ACC_A$
& $FPR_L$ & $FPR_A$ & $AUC_L$ & $AUC_A$ & $ACC_L$ & $ACC_A$ \\
\midrule

MLP-Head
& 68.40	& 55.55	& 77.71	& 83.46	& 70.79	& 81.25
& 83.77	& 66.37	& 67.78	& 77.14	& 53.46	& 69.71
& 80.20	& 70.34	& 67.98	& 72.57	& 40.21	& 60.52
& 77.46	& 64.09	& 71.16	& 77.72	& 54.82	& 70.49 \\

Linear-Head 
& 66.16 & 54.13 & 79.30 & 84.46 & 73.85 & 82.97
& 71.93 & 60.00 & 77.41 & 82.45 & 68.35 & 80.51
& 67.91 & 63.37 & 79.56 & 80.28 & 61.69 & 74.78
& 68.67 & 59.17 & 78.76 & 82.40 & 67.96 & 79.42 \\

Linear-Head+Guidance
& 62.22	& 58.64	& 79.20	& 83.78	& 73.85	& 82.97
& 72.08	& 62.27	& 77.67	& 82.40	& 68.35	& 80.51
& 66.78	& 61.51	& 80.32	& 81.45	& 61.69	& 74.78
& 67.03	& 60.81	& 79.06	& 82.54	& 67.96	& 79.42 \\

\midrule
KAN B-spline	
& 67.15	& 60.96	& 78.17	& 81.20	& 68.73	& 77.80
& 75.37	& 68.04	& 73.55	& 77.51	& 59.13	& 70.47
& 68.82	& 65.00	& 78.65	& 81.28	& 53.55	& 67.39
& 70.45	& 64.67	& 76.79	& 80.00	& 60.47	& 71.89 \\

KAN B-spline+Guidance
& 65.56	& 61.82	& 78.78	& 81.76	& 68.73	& 77.80
& 71.67	& 66.56	& 74.48	& 78.42	& 59.13	& 70.47
& 68.55	& 64.14	& 79.26	& 81.98	& 53.55	& 67.39
& 68.59	& 64.17	& 77.51	& 80.72	& 60.47	& 71.89 \\

\midrule
KAN-Head(Baseline) 
& 59.65 & 53.84 & 83.48 & 86.37 & \textbf{81.55} & \textbf{87.18}
& 71.11 & 57.77 & 80.53 & 84.89 & \textbf{78.29} & \textbf{85.87}
& 59.69 & 56.38 & 81.86 & 84.22 & 75.24 & 83.05
& 63.48 & 56.00 & 81.96 & 85.16 & \textbf{78.36} & \textbf{85.37} \\

KAN-Head $\alpha_{\min}$ 
& 57.85 & 53.44 & 84.23 & 86.89 & 80.78 & 86.67
& 68.51 & 56.23 & 80.46 & 85.18 & 73.95 & 83.55
& 54.01 & {53.17} & 82.11 & \textbf{85.33} & 73.60 & 83.35
& 60.12 & 54.28 & 82.27 & 85.80 & 76.11 & 84.52 \\

KAN-Head trainable $\sigma$ 
& 60.29 & 54.41 & 82.97 & 86.15 & 81.10 & 87.00
& 71.52 & 58.00 & 80.31 & 84.77 & 78.22 & 85.85
& 62.27 & 56.49 & 81.31 & 84.21 & 75.15 & 83.22
& 64.69 & 56.30 & 81.53 & 85.04 & 78.16 & 85.36 \\

\midrule
KAN-Head+Guidance 
& 54.64 & 51.89 & 84.18 & 86.80 & {81.55} & {87.18}
& \textbf{63.20} & \textbf{54.28} & 81.59 & \textbf{85.60} & {78.29} & {85.87}
& 57.53 & 56.21 & 82.49 & 85.08 & 75.24 & 83.05
& 58.46 & 54.13 & 82.75 & 85.83 & {78.36} & {85.37} \\

\midrule
TA-KAN Head
& 52.11 & 52.19 & 85.86 & 87.53 & 80.86 & 86.80
& 70.63 & 58.31 & 81.55 & 84.92 & 77.49 & 85.10
& 51.35 & 54.98 & 84.42 & 84.48 & {75.70} & {83.82}
& 58.03 & 55.16 & 83.94 & 85.64 & 78.02 & 85.24 \\

TA-KAN+Guidance(ours) 
& \textbf{48.73} & \textbf{48.52} & \textbf{86.42} & \textbf{88.06} & 80.86 & 86.80
& 68.28 & 54.94 & \textbf{81.99} & 85.51 & 77.49 & 85.10
& \textbf{50.48} & \textbf{53.07} & \textbf{84.77} & 84.96 & \textbf{75.70} & \textbf{83.82}
& \textbf{55.83} & \textbf{52.18} & \textbf{84.39} & \textbf{86.18} & 78.02 & 85.24 \\

\bottomrule
\end{tabular}
} 
}
\label{tab:ablation_detailed}
\end{table*}
We present detailed ablation results in \cref{tab:ablation_detailed}, analyzing the contribution of each component of our method across all task settings.

\textbf{Comparison with B-spline KAN.}
In addition to the ablations of different components presented in \cref{tab:ablation_summary} of the main paper, we evaluate the effect of the RBF basis by comparing with the original cubic B-spline KAN \cite{KAN,Bspline} (\textit{KAN B-spline} in \cref{tab:ablation_detailed}) under the same experimental setting, using a capacity-comparable configuration with $G=13$ grid points and spline order $k=3$.
The B-spline KAN underperforms the fixed-width RBF-KAN (\textit{KAN-Head}) in OOD detection, while applying guidance improves both variants.
These results support the use of RBF bases for OOD-CL, consistent with their advantages in high-dimensional settings~\cite{RBFKAN2,RBFKAN,KAC}.
Importantly, TA-KAN further improves upon the fixed-width RBF-KAN, isolating the additional benefit of the proposed task-adaptive locality mechanism.

\subsection{Hyperparameter Analysis} \label{app:ablation_hyperpara}

\begin{table}[t]
\centering
\caption{Results on CUB-200 dataset for different $\alpha_{\min}$ values. Lower FPR is better; higher AUC and ACC are better.}
\small
\renewcommand{\arraystretch}{1.1}
\setlength{\tabcolsep}{4pt}
\resizebox{0.6\textwidth}{!}{
\begin{tabular}{l|cccccc}
\toprule
$\alpha_{\min}$ 
& $FPR_L$ & $FPR_A$ & $AUC_L$ & $AUC_A$ & $ACC_L$ & $ACC_A$ \\
\midrule
1     & 59.65 & 53.84 & 83.48 & 86.37 & 81.55 & 87.18 \\
1  + Guidance   & 54.64 & 51.89 & 84.18 & 86.80 & 81.55 & 87.18 \\
\midrule
0.75  & 55.78 & 53.11 & 84.61 & 86.90 & 81.47 & 87.10 \\
0.75 + Guidance & 51.46 & 49.40 & 85.28 & 87.42 & 81.47 & 87.10 \\
\midrule
0.5   & 52.11 & 52.19 & 85.86 & 87.53 & 80.86 & 86.80 \\
0.5 + Guidance  & {48.73} & {48.52} & {86.42} & {88.06} & 80.86 & 86.80 \\
\midrule
0.25  & 50.22 & 52.87 & 85.82 & 86.93 & 78.43 & 84.57 \\
0.25 + Guidance & 49.13 & 50.96 & 86.15 & 87.44 & 78.43 & 84.57 \\
\bottomrule
\end{tabular}
}
\label{tab:alphamin}
\end{table}

\textbf{Effect of $\alpha_{\min}$.}
We analyze the impact of the hyperparameter $\alpha_{\min}$ in TA-KAN in \cref{tab:alphamin}, considering values from $\{0.25, 0.5, 0.75\}$, where $\alpha_{\min}=1$ corresponds to the standard KAN without adaptive locality.
We observe that moderate values of $\alpha_{\min}$ yield the best overall performance.
As $\alpha_{\min}$ decreases, the RBF responses become increasingly localized, which initially improves OOD detection by enhancing fine-grained separability between ID and OOD samples.
However, excessive reduction in $\alpha_{\min}$ leads to overly sharp responses, making the model sensitive to intra-class variations and thereby degrading both OOD detection gains and ID classification performance.
This behavior highlights an inherent trade-off between locality and generalizability: while increased locality improves discrimination in ambiguous regions, sufficient smoothness is required to preserve robustness to intra-class variability.


\textbf{Assumptions on $N$.}
We assume the number of tasks $N$ is known when defining the schedule in \cref{eq:alphat}, following standard practice in class-incremental and prompt-based continual learning methods \cite{Dual_prompt, CODA_prompt, CPrompt}.
While we consider such a known (or bounded) set of tasks, this does not limit practical applicability, as new tasks can be accommodated by incrementally introducing additional task prototypes, with the corresponding width factor initialized to 1 (i.e., the default, most general setting).

\textbf{Threshold selection in continual learning.}
In practical deployment, OOD detection requires selecting a decision threshold. 
Following prior work, a practical approach is to estimate the threshold using statistics of the current task data (e.g., average-based thresholding), or to use a small held-out validation set when available~\cite{MORE1,proKT}. 
Our method is fully compatible with such strategies at deployment time.
For evaluation, following prior work~\cite{proKT,ORE,MORE1,OODCILunsuper}, we do not perform explicit threshold tuning and instead adopt threshold-independent metrics, namely AUC and FPR, which enable fair and consistent comparison across methods.

\subsection{Comparison with Various Post-Hoc OOD Scores} \label{app:ablation_posthoc}
\begin{table*}[tbh!]
\centering
\caption{Results on the CUB-200~\cite{CUB} dataset using various post-hoc OOD scoring methods across task settings. Lower FPR is better; higher AUC and ACC are better. Values in parentheses denote changes relative to the corresponding Linear-Head scores.}
\small
\renewcommand{\arraystretch}{1.1}
\setlength{\tabcolsep}{4pt}
\resizebox{0.98\textwidth}{!}{
\begin{tabular}{llccc|ccc|ccc}
\toprule
\multirow{2}{*}{\textbf{Method}} & \multirow{2}{*}{\textbf{Score}} 
& \multicolumn{3}{c|}{\textbf{5 tasks}} 
& \multicolumn{3}{c|}{\textbf{10 tasks}} 
& \multicolumn{3}{c}{\textbf{20 tasks}} \\
\cmidrule(lr){3-5} \cmidrule(lr){6-8} \cmidrule(lr){9-11}
& & $FPR_L$ & $AUC_L$ & $ACC_L$ 
& $FPR_L$ & $AUC_L$ & $ACC_L$ 
& $FPR_L$ & $AUC_L$ & $ACC_L$ \\
\midrule

\multirow{6}{*}{Linear-Head}
& MSP~\cite{hendrycks_2017MSP}       & 66.16 & 79.30 & \multirow{6}{*}{73.85} & 71.93 & 77.41 & \multirow{6}{*}{68.35} & 67.91 & 79.56 & \multirow{6}{*}{61.69} \\
& MaxLogit~\cite{MaxLogit}  & 59.36 & 84.16 &  & 66.41 & 80.91 &  & 70.71 & 78.08 &  \\
& Energy~\cite{EnergyOE}    & 59.66 & 83.75 &  & 66.13 & 79.90 &  & 76.89 & 73.42 &  \\
& Entropy~\cite{MetaOOD}   & 64.43 & 82.40 &  & 68.34 & 81.49 &  & 65.02 & 80.63 &  \\
& KL~\cite{MaxLogit}        & 60.53 & 82.89 &  & 67.56 & 78.73 &  & 78.43 & 72.71 &  \\
& MSP-BC-CE~\cite{OODCILunsuper}  & 64.70 & 80.36 &  & 67.97 & 78.29 &  & 74.28 & 76.04 &  \\

\midrule
\multirow{6}{*}{KAN-Head (Baseline)}
& MSP~\cite{hendrycks_2017MSP}
& 59.65 {\scriptsize \textcolor{blue}{-6.51}}
& 83.48 {\scriptsize \textcolor{blue}{+4.18}}
& \multirow{6}{*}{81.55}
& 71.11 {\scriptsize \textcolor{blue}{-0.82}}
& 80.53 {\scriptsize \textcolor{blue}{+3.12}}
& \multirow{6}{*}{78.29}
& 59.69 {\scriptsize \textcolor{blue}{-8.22}}
& 81.86 {\scriptsize \textcolor{blue}{+2.30}}
& \multirow{6}{*}{75.24} \\

& MaxLogit~\cite{MaxLogit}
& 54.17 {\scriptsize \textcolor{blue}{-5.19}}
& 85.38 {\scriptsize \textcolor{blue}{+1.22}}
&
& 56.97 {\scriptsize \textcolor{blue}{-9.44}}
& 82.70 {\scriptsize \textcolor{blue}{+1.79}}
&
& 59.65 {\scriptsize \textcolor{blue}{-11.06}}
& 80.67 {\scriptsize \textcolor{blue}{+2.59}} \\

& Energy~\cite{EnergyOE}
& 56.57 {\scriptsize \textcolor{blue}{-3.09}}
& 84.47 {\scriptsize \textcolor{blue}{+0.72}}
& 
& 57.15 {\scriptsize \textcolor{blue}{-8.98}}
& 81.68 {\scriptsize \textcolor{blue}{+1.78}}
&
& 63.47 {\scriptsize \textcolor{blue}{-13.42}}
& 78.30 {\scriptsize \textcolor{blue}{+4.88}} \\

& Entropy~\cite{MetaOOD}
& 55.38 {\scriptsize \textcolor{blue}{-9.05}}
& 85.20 {\scriptsize \textcolor{blue}{+2.80}}
&
& 66.39 {\scriptsize \textcolor{blue}{-1.95}}
& 82.41 {\scriptsize \textcolor{blue}{+0.92}}
&
& 54.84 {\scriptsize \textcolor{blue}{-10.18}}
& 82.33 {\scriptsize \textcolor{blue}{+1.70}} \\

& KL~\cite{MaxLogit}
& 58.53 {\scriptsize \textcolor{blue}{-2.00}}
& 84.04 {\scriptsize \textcolor{blue}{+1.15}}
&
& 58.16 {\scriptsize \textcolor{blue}{-9.40}}
& 81.42 {\scriptsize \textcolor{blue}{+2.69}}
&
& 63.30 {\scriptsize \textcolor{blue}{-15.13}}
& 78.37 {\scriptsize \textcolor{blue}{+5.66}} \\

& MSP-BC-CE~\cite{OODCILunsuper} 
& 58.01 {\scriptsize \textcolor{blue}{-6.69}}
& 84.60 {\scriptsize \textcolor{blue}{+4.24}}
&
& 67.88 {\scriptsize \textcolor{blue}{-0.09}}
& 80.66 {\scriptsize \textcolor{blue}{+2.37}}
&
& 60.65 {\scriptsize \textcolor{blue}{-13.63}}
& 80.20 {\scriptsize \textcolor{blue}{+4.16}} \\

\midrule
\multirow{6}{*}{TA-KAN Head}
& MSP~\cite{hendrycks_2017MSP}
& 52.11 {\scriptsize \textcolor{blue}{-14.05}}
& 85.86 {\scriptsize \textcolor{blue}{+6.56}}
& \multirow{6}{*}{80.86}
& 70.63 {\scriptsize \textcolor{blue}{-1.30}}
& 81.55 {\scriptsize \textcolor{blue}{+4.14}}
& \multirow{6}{*}{77.49}
& 51.35 {\scriptsize \textcolor{blue}{-16.56}}
& 84.42 {\scriptsize \textcolor{blue}{+4.86}}
& \multirow{6}{*}{75.70} \\

& MaxLogit~\cite{MaxLogit}
& 52.91 {\scriptsize \textcolor{blue}{-6.45}}
& 85.77 {\scriptsize \textcolor{blue}{+1.61}}
&
& 57.75 {\scriptsize \textcolor{blue}{-8.66}}
& 82.25 {\scriptsize \textcolor{blue}{+1.34}}
&
& 55.34 {\scriptsize \textcolor{blue}{-15.37}}
& 82.17 {\scriptsize \textcolor{blue}{+4.09}} \\

& Energy~\cite{EnergyOE}
& 58.69 {\scriptsize \textcolor{blue}{-0.97}}
& 84.15 {\scriptsize \textcolor{blue}{+0.40}}
&
& 61.86 {\scriptsize \textcolor{blue}{-4.27}}
& 80.35 {\scriptsize \textcolor{blue}{+0.45}}
&
& 62.36 {\scriptsize \textcolor{blue}{-14.53}}
& 77.65 {\scriptsize \textcolor{blue}{+4.23}} \\

& Entropy~\cite{MetaOOD}
& 52.22 {\scriptsize \textcolor{blue}{-12.21}}
& 85.77 {\scriptsize \textcolor{blue}{+3.37}}
&
& 66.49 {\scriptsize \textcolor{blue}{-1.85}}
& 81.62 {\scriptsize \textcolor{blue}{+0.13}}
&
& 57.27 {\scriptsize \textcolor{blue}{-7.75}}
& 81.35 {\scriptsize \textcolor{blue}{+0.72}} \\

& KL~\cite{MaxLogit}
& 58.25 {\scriptsize \textcolor{blue}{-2.28}}
& 83.67 {\scriptsize \textcolor{blue}{+0.78}}
&
& 67.13 {\scriptsize \textcolor{blue}{-0.43}}
& 77.57 {\scriptsize \textcolor{red}{-1.16}}
&
& 66.18 {\scriptsize \textcolor{blue}{-12.25}}
& 76.55 {\scriptsize \textcolor{blue}{+3.84}} \\

& MSP-BC-CE~\cite{OODCILunsuper} 
& 54.27 {\scriptsize \textcolor{blue}{-10.43}}
& 85.24 {\scriptsize \textcolor{blue}{+4.88}}
&
& 67.96 {\scriptsize \textcolor{blue}{-0.01}}
& 80.32 {\scriptsize \textcolor{blue}{+2.03}}
&
& 58.29 {\scriptsize \textcolor{blue}{-15.99}}
& 81.86 {\scriptsize \textcolor{blue}{+5.82}} \\

\midrule
\multirow{6}{*}{TA-KAN+Guidance}
& MSP~\cite{hendrycks_2017MSP}
& 48.73 {\scriptsize \textcolor{blue}{-17.43}}
& 86.42 {\scriptsize \textcolor{blue}{+7.12}}
& \multirow{6}{*}{80.86}
& 68.28 {\scriptsize \textcolor{blue}{-3.65}}
& 81.99 {\scriptsize \textcolor{blue}{+4.58}}
& \multirow{6}{*}{77.49}
& 50.48 {\scriptsize \textcolor{blue}{-17.43}}
& 84.77 {\scriptsize \textcolor{blue}{+5.21}}
& \multirow{6}{*}{75.70} \\

& MaxLogit~\cite{MaxLogit}
& 50.51 {\scriptsize \textcolor{blue}{-8.85}}
& 86.03 {\scriptsize \textcolor{blue}{+1.87}}
&
& 58.37 {\scriptsize \textcolor{blue}{-8.04}}
& 82.38 {\scriptsize \textcolor{blue}{+1.47}}
&
& 57.57 {\scriptsize \textcolor{blue}{-13.14}}
& 82.07 {\scriptsize \textcolor{blue}{+3.99}} \\

& Energy~\cite{EnergyOE}
& 58.32 {\scriptsize \textcolor{blue}{-1.34}}
& 84.71 {\scriptsize \textcolor{blue}{+0.96}}
&
& 68.17 {\scriptsize \textcolor{red}{+2.04}}
& 79.56 {\scriptsize \textcolor{red}{-0.34}}
&
& 67.00 {\scriptsize \textcolor{blue}{-9.89}}
& 76.65 {\scriptsize \textcolor{blue}{+3.23}} \\

& Entropy~\cite{MetaOOD}
& 52.94 {\scriptsize \textcolor{blue}{-11.49}}
& 85.24 {\scriptsize \textcolor{blue}{+2.84}}
&
& 67.25 {\scriptsize \textcolor{blue}{-1.09}}
& 81.97 {\scriptsize \textcolor{blue}{+0.48}}
&
& 57.31 {\scriptsize \textcolor{blue}{-7.71}}
& 79.85 {\scriptsize \textcolor{red}{-0.78}} \\

& KL~\cite{MaxLogit}
& 56.37 {\scriptsize \textcolor{blue}{-4.16}}
& 83.90 {\scriptsize \textcolor{blue}{+1.01}}
&
& 69.26 {\scriptsize \textcolor{red}{+1.70}}
& 77.59 {\scriptsize \textcolor{red}{-1.14}}
&
& 67.47 {\scriptsize \textcolor{blue}{-10.96}}
& 76.49 {\scriptsize \textcolor{blue}{+3.78}} \\

& MSP-BC-CE~\cite{OODCILunsuper} 
& 52.92 {\scriptsize \textcolor{blue}{-11.78}}
& 85.50 {\scriptsize \textcolor{blue}{+5.14}}
&
& 67.67 {\scriptsize \textcolor{blue}{-0.30}}
& 80.54 {\scriptsize \textcolor{blue}{+2.25}}
&
& 58.38 {\scriptsize \textcolor{blue}{-15.90}}
& 81.99 {\scriptsize \textcolor{blue}{+5.95}} \\

\bottomrule
\end{tabular}
}
\label{tab:cub_posthoc}
\end{table*}

\begin{table*}[tbh!]
\centering
\caption{Results on the ImageNet-R~\cite{ImageNet-R} dataset using various post-hoc OOD scoring methods across task settings. Lower FPR is better; higher AUC and ACC are better. Values in parentheses denote changes relative to the corresponding Linear-Head scores.}
\small
\renewcommand{\arraystretch}{1.1}
\setlength{\tabcolsep}{4pt}
\resizebox{0.98\textwidth}{!}{
\begin{tabular}{llccc|ccc|ccc}
\toprule
\multirow{2}{*}{\textbf{Method}} & \multirow{2}{*}{\textbf{Score}} 
& \multicolumn{3}{c|}{\textbf{5 tasks}} 
& \multicolumn{3}{c|}{\textbf{10 tasks}} 
& \multicolumn{3}{c}{\textbf{20 tasks}} \\
\cmidrule(lr){3-5} \cmidrule(lr){6-8} \cmidrule(lr){9-11}
& & $FPR_L$ & $AUC_L$ & $ACC_L$ 
& $FPR_L$ & $AUC_L$ & $ACC_L$ 
& $FPR_L$ & $AUC_L$ & $ACC_L$ \\
\midrule

\multirow{6}{*}{Linear-Head}
& MSP~\cite{hendrycks_2017MSP}       & 42.74 & 85.44 & \multirow{6}{*}{77.96} & 45.69 & 86.26 & \multirow{6}{*}{76.51} & 53.49 & 85.28 & \multirow{6}{*}{73.90} \\
& MaxLogit~\cite{MaxLogit}  & 47.13 & 87.36 &  & 53.63 & 87.09 &  & 63.64 & 84.24 &  \\
& Energy~\cite{EnergyOE}    & 48.04 & 87.27 &  & 55.12 & 86.43 &  & 66.65 & 82.06 &  \\
& Entropy~\cite{MetaOOD}   & 41.98 & 87.21 &  & 44.95 & 87.95 &  & 52.62 & 86.52 &  \\
& KL~\cite{MaxLogit}        & 52.69 & 86.51 &  & 59.11 & 84.98 &  & 68.95 & 80.76 &  \\
& MSP-BC-CE~\cite{OODCILunsuper}  & 43.37 & 86.39 &  & 47.26 & 86.46 &  & 55.62 & 84.99 &  \\

\midrule
\multirow{6}{*}{KAN-Head (Baseline)}
& MSP~\cite{hendrycks_2017MSP} & 42.52 {\scriptsize \textcolor{blue}{-0.22}} & 84.81 {\scriptsize \textcolor{red}{-0.63}} & \multirow{6}{*}{79.25}
& 43.04 {\scriptsize \textcolor{blue}{-2.65}} & 85.64 {\scriptsize \textcolor{red}{-0.62}} & \multirow{6}{*}{78.76}
& 44.00 {\scriptsize \textcolor{blue}{-9.49}} & 85.40 {\scriptsize \textcolor{blue}{+0.12}} & \multirow{6}{*}{77.42} \\
& MaxLogit~\cite{MaxLogit} & 39.76 {\scriptsize \textcolor{blue}{-7.37}} & 87.29 {\scriptsize \textcolor{red}{-0.07}} &
& 40.37 {\scriptsize \textcolor{blue}{-13.26}} & 87.47 {\scriptsize \textcolor{blue}{+0.38}} &
& 41.88 {\scriptsize \textcolor{blue}{-21.76}} & 86.79 {\scriptsize \textcolor{blue}{+2.55}} & \\
& Energy~\cite{EnergyOE} & 39.99 {\scriptsize \textcolor{blue}{-8.05}} & 87.54 {\scriptsize \textcolor{blue}{+0.27}} & 
& 40.60 {\scriptsize \textcolor{blue}{-14.52}} & 87.65 {\scriptsize \textcolor{blue}{+1.22}} &
& 42.47 {\scriptsize \textcolor{blue}{-24.18}} & 86.92 {\scriptsize \textcolor{blue}{+4.86}} & \\
& Entropy~\cite{MetaOOD} & 41.84 {\scriptsize \textcolor{blue}{-0.14}} & 85.88 {\scriptsize \textcolor{red}{-1.33}} &
& 41.75 {\scriptsize \textcolor{blue}{-3.20}} & 86.82 {\scriptsize \textcolor{red}{-1.13}} &
& 42.98 {\scriptsize \textcolor{blue}{-9.64}} & 86.44 {\scriptsize \textcolor{red}{-0.08}} & \\
& KL~\cite{MaxLogit} & 40.85 {\scriptsize \textcolor{blue}{-11.84}} & 86.97 {\scriptsize \textcolor{blue}{+0.46}} &
& 40.89 {\scriptsize \textcolor{blue}{-18.22}} & 87.60 {\scriptsize \textcolor{blue}{+2.62}} &
& 42.47 {\scriptsize \textcolor{blue}{-26.48}} & 87.01 {\scriptsize \textcolor{blue}{+6.25}} & \\
& MSP-BC-CE~\cite{OODCILunsuper}  & 40.65 {\scriptsize \textcolor{blue}{-2.72}} & 85.50 {\scriptsize \textcolor{red}{-0.89}} &
& 42.12 {\scriptsize \textcolor{blue}{-5.14}} & 86.84 {\scriptsize \textcolor{blue}{+0.38}} &
& 44.88 {\scriptsize \textcolor{blue}{-10.74}} & 85.78 {\scriptsize \textcolor{blue}{+0.79}} & \\

\midrule
\multirow{6}{*}{TA-KAN Head}
& MSP~\cite{hendrycks_2017MSP} & 40.07 {\scriptsize \textcolor{blue}{-2.67}} & 85.77 {\scriptsize \textcolor{blue}{+0.33}} & \multirow{6}{*}{78.83}
& 40.25 {\scriptsize \textcolor{blue}{-5.44}} & 86.76 {\scriptsize \textcolor{blue}{+0.50}} & \multirow{6}{*}{77.76}
& 42.57 {\scriptsize \textcolor{blue}{-10.92}} & 85.99 {\scriptsize \textcolor{blue}{+0.71}} & \multirow{6}{*}{77.03} \\
& MaxLogit~\cite{MaxLogit} & 38.64 {\scriptsize \textcolor{blue}{-8.49}} & 87.37 {\scriptsize \textcolor{blue}{+0.01}} &
& 38.87 {\scriptsize \textcolor{blue}{-14.76}} & 87.67 {\scriptsize \textcolor{blue}{+0.58}} &
& 42.06 {\scriptsize \textcolor{blue}{-21.58}} & 86.76 {\scriptsize \textcolor{blue}{+2.52}} & \\
& Energy~\cite{EnergyOE} & 39.51 {\scriptsize \textcolor{blue}{-8.53}} & 87.36 {\scriptsize \textcolor{blue}{+0.09}} &
& 40.37 {\scriptsize \textcolor{blue}{-14.75}} & 87.00 {\scriptsize \textcolor{blue}{+0.57}} &
& 45.43 {\scriptsize \textcolor{blue}{-21.22}} & 85.62 {\scriptsize \textcolor{blue}{+3.56}} & \\
& Entropy~\cite{MetaOOD} & 39.57 {\scriptsize \textcolor{blue}{-2.41}} & 87.52 {\scriptsize \textcolor{blue}{+0.31}} &
& 39.60 {\scriptsize \textcolor{blue}{-5.35}} & 87.98 {\scriptsize \textcolor{blue}{+0.03}} &
& 42.76 {\scriptsize \textcolor{blue}{-9.86}} & 86.87 {\scriptsize \textcolor{blue}{+0.35}} & \\
& KL~\cite{MaxLogit} & 40.47 {\scriptsize \textcolor{blue}{-12.22}} & 86.58 {\scriptsize \textcolor{blue}{+0.07}} &
& 39.87 {\scriptsize \textcolor{blue}{-19.24}} & 87.76 {\scriptsize \textcolor{blue}{+2.78}} &
& 43.99 {\scriptsize \textcolor{blue}{-24.96}} & 86.69 {\scriptsize \textcolor{blue}{+5.93}} & \\
& MSP-BC-CE~\cite{OODCILunsuper}  & 40.27 {\scriptsize \textcolor{blue}{-3.10}} & 86.70 {\scriptsize \textcolor{blue}{+0.31}} &
& 40.48 {\scriptsize \textcolor{blue}{-6.78}} & 87.22 {\scriptsize \textcolor{blue}{+0.76}} &
& 43.84 {\scriptsize \textcolor{blue}{-11.78}} & 85.99 {\scriptsize \textcolor{blue}{+1.00}} & \\

\midrule
\multirow{6}{*}{TA-KAN+Guidance}
& MSP~\cite{hendrycks_2017MSP} & 39.72 {\scriptsize \textcolor{blue}{-3.02}} & 85.92 {\scriptsize \textcolor{blue}{+0.48}} & \multirow{6}{*}{78.83}
& 39.98 {\scriptsize \textcolor{blue}{-5.71}} & 86.87 {\scriptsize \textcolor{blue}{+0.61}} & \multirow{6}{*}{77.76}
& 42.61 {\scriptsize \textcolor{blue}{-10.88}} & 86.10 {\scriptsize \textcolor{blue}{+0.82}} & \multirow{6}{*}{77.03} \\
& MaxLogit~\cite{MaxLogit} & 38.64 {\scriptsize \textcolor{blue}{-8.49}} & 87.58 {\scriptsize \textcolor{blue}{+0.22}} &
& 39.03 {\scriptsize \textcolor{blue}{-14.60}} & 87.64 {\scriptsize \textcolor{blue}{+0.55}} &
& 42.82 {\scriptsize \textcolor{blue}{-20.82}} & 86.61 {\scriptsize \textcolor{blue}{+2.37}} & \\
& Energy~\cite{EnergyOE} & 39.77 {\scriptsize \textcolor{blue}{-8.27}} & 87.29 {\scriptsize \textcolor{blue}{+0.02}} &
& 43.15 {\scriptsize \textcolor{blue}{-11.97}} & 85.72 {\scriptsize \textcolor{red}{-0.71}} &
& 56.82 {\scriptsize \textcolor{blue}{-9.83}} & 81.81 {\scriptsize \textcolor{red}{-0.25}} & \\
& Entropy~\cite{MetaOOD} & 39.99 {\scriptsize \textcolor{blue}{-1.99}} & 87.24 {\scriptsize \textcolor{blue}{+0.03}} &
& 40.14 {\scriptsize \textcolor{blue}{-4.81}} & 87.41 {\scriptsize \textcolor{red}{-0.54}} &
& 44.14 {\scriptsize \textcolor{blue}{-8.48}} & 86.63 {\scriptsize \textcolor{blue}{+0.11}} & \\
& KL~\cite{MaxLogit} & 39.79 {\scriptsize \textcolor{blue}{-12.90}} & 86.91 {\scriptsize \textcolor{blue}{+0.40}} &
& 39.91 {\scriptsize \textcolor{blue}{-19.20}} & 87.61 {\scriptsize \textcolor{blue}{+2.63}} &
& 45.10 {\scriptsize \textcolor{blue}{-23.85}} & 86.16 {\scriptsize \textcolor{blue}{+5.40}} & \\
& MSP-BC-CE~\cite{OODCILunsuper}  & 39.61 {\scriptsize \textcolor{blue}{-3.76}} & 87.14 {\scriptsize \textcolor{blue}{+0.75}} &
& 40.35 {\scriptsize \textcolor{blue}{-6.91}} & 87.35 {\scriptsize \textcolor{blue}{+0.89}} &
& 43.78 {\scriptsize \textcolor{blue}{-11.84}} & 86.05 {\scriptsize \textcolor{blue}{+1.06}} & \\

\bottomrule
\end{tabular}
}
\label{tab:imagenetr_posthoc}
\end{table*}

We compare our method with standard post-hoc OOD scoring methods on CUB-200 (\cref{tab:cub_posthoc}) and ImageNet-R (\cref{tab:imagenetr_posthoc}), including MSP~\cite{hendrycks_2017MSP}, MaxLogit~\cite{MaxLogit}, Energy~\cite{EnergyOE}, Entropy~\cite{MetaOOD}, and KL~\cite{MaxLogit}. 
Across both datasets, we observe a consistent trend similar to MSP: replacing the standard KAN with the proposed TA-KAN consistently improves OOD detection performance in terms of both FPR and AUC, while preserving the ID performance gains over the linear head, across all considered confidence-based scores. 
Furthermore, incorporating the proposed guidance into TA-KAN yields additional improvements.

\subsection{Comparison with Geometry-Based OOD Scores} \label{app:ablation_geometry}
\begin{table}[t]
\centering
\caption{Comparison of different geometry-based scoring methods on the CUB-200 dataset. Lower FPR is better; higher AUC and ACC are better.}
\small
\setlength{\tabcolsep}{4pt}
\renewcommand{\arraystretch}{1.1}
\resizebox{0.7\textwidth}{!}{
\begin{tabular}{lcccccc}
\toprule
Method 
& $FPR_L$
& $FPR_A$
& $AUC_L$
& $AUC_A$
& $ACC_L$
& $ACC_A$ \\
\midrule

MSP & 53.50 & 50.80 & 85.70 & 87.61 & 81.41 & 86.97 \\
KNN & 85.33 & 71.43 & 57.16 & 69.47 & 81.41 & 86.97 \\
NNGuide & 55.16 & 51.69 & 86.01 & 87.89 & 81.41 & 86.97 \\
Guidance (Ours) & \textbf{50.30} & \textbf{47.21} & \textbf{86.20} & \textbf{88.15} & \textbf{81.41} & \textbf{86.97} \\

\bottomrule
\end{tabular}
}
\label{tab:geometry_ablation}
\end{table}
We compare the proposed geometry-guided OOD score with existing geometry-based OOD methods that operate in the embedding space, as shown in \cref{tab:geometry_ablation}. 
These methods typically require access to past ID data; therefore, for a fair comparison, we equip them with a rehearsal buffer of size 1000.
Our proposed guidance consistently outperforms these approaches. 
In particular, NNGuide combines embedding-space geometry with classifier confidence (e.g., MSP), whereas our method provides geometry guidance in the task (prompt) space without requiring access to past ID data. 
Despite operating in a rehearsal-free setting, our approach achieves significantly better performance, highlighting the effectiveness of proposed geometry guided score for OOD-CL.

\subsection{Backbone and Prompt-Based Model} \label{app:ablation_prompt}
\begin{table}[t]
\centering
\caption{Results on the CUB-200 and ImageNet-R datasets on 10-tasks setting using DualPrompt as the backbone prompt-based model. Lower FPR is better; higher AUC and ACC are better.}
\small
\setlength{\tabcolsep}{4pt}
\renewcommand{\arraystretch}{1.1}
\resizebox{0.7\textwidth}{!}{
\begin{tabular}{lccc|ccc}
\toprule
\multirow{2}{*}{Method} 
& \multicolumn{3}{c|}{CUB-200} 
& \multicolumn{3}{c}{ImageNet-R} \\
\cmidrule(lr){2-4} \cmidrule(lr){5-7}
& $FPR_L$ & $AUC_L$ & $ACC_L$ 
& $FPR_L$ & $AUC_L$ & $ACC_L$ \\
\midrule

Linear-Head 
& 79.15 & 76.60 & 70.43
& 50.84 & 83.16 & 72.62 \\

KAN-Head (Baseline) 
& 71.81 & 80.16 & 79.28
& 46.93 & 83.30 & 74.34 \\

TA-KAN 
& 67.09 & \textbf{81.86} & 79.51
& 46.43 & 84.12 & 75.10 \\

TA-KAN+Guidance 
& \textbf{62.83} & 80.67 & \textbf{79.51}
& \textbf{44.98} & \textbf{84.80} & \textbf{75.10} \\

\bottomrule
\end{tabular}
}
\label{tab:backbone_ablation}
\end{table}
We adopt a prompt-based continual learning model with a frozen backbone $g$ (e.g., ViT) as the feature extractor, following prior work~\cite{KAC,CODA_prompt,CPrompt,Dual_prompt}. 
Our proposed method is orthogonal to the choice of backbone and prompt-learning strategy. 
Therefore, it can be seamlessly integrated with different prompt-based CIL frameworks without modifying the backbone or prompt-learning procedure. 
In our main experiments, we use CODAPrompt~\cite{CODA_prompt} as the default model, while here we evaluate generalization using DualPrompt~\cite{Dual_prompt}.
As shown in \cref{tab:backbone_ablation}, we observe a consistent trend across models. 
Replacing the standard KAN head with TA-KAN improves OOD detection performance and ID performance. 
Further incorporating the proposed geometry-based guidance yields additional improvements.

\subsection{Efficiency Analysis} \label{app:efficiency}

We further assess the performance--efficiency trade-off by measuring the average training time per epoch and inference latency per sample under identical experimental settings on an NVIDIA A100 GPU.
As shown in \cref{tab:efficiency}, TA-KAN increases training time by only $0.67\%$ and inference latency by $0.50\%$ relative to the Linear-Head, while substantially reducing FPR from $66.16\%$ to $52.11\%$ (a $21.2\%$ relative reduction) and improving AUC from $79.30\%$ to $85.86\%$.
This small runtime overhead is expected because all methods share the same ViT backbone and prompt-learning module, which dominate the overall computation, whereas the architectural differences are confined to the lightweight classifier head.
Importantly, TA-KAN introduces no additional task-specific classifier parameters; instead, it maintains the same number of learnable parameters as the standard KAN-Head while adaptively modulating the RBF width factor, with nearly identical runtime.
The proposed geometry-guided OOD score is also computationally efficient, as it is computed post hoc using prompt alignment.
Since the size of the prompt bank is much smaller than the number of training samples, this incurs substantially lower cost than conventional geometry-based OOD methods such as KNN or NNGuide, which require nearest-neighbor search over stored ID features.

\begin{table}[tbh!]
\centering
\caption{Performance--efficiency comparison on the CUB-200 5-task setting.}
\label{tab:efficiency}
\small
\setlength{\tabcolsep}{2pt}
\resizebox{0.85\textwidth}{!}{
\begin{tabular}{lcccc}
\toprule
\textbf{Method} & \textbf{$FPR_L\downarrow$} & \textbf{$AUC_L \uparrow$} &
\textbf{Train Time / Epoch (s) $\downarrow$} &
\textbf{Inference Time / Sample (ms) $\downarrow$} \\
\midrule
Linear-Head & 66.16 & 79.30 & \textbf{11.6062} & \textbf{4.5659} \\
KAN-Head    & 59.65 & 83.48 & 11.6449 & 4.5895 \\
TA-KAN      & \textbf{52.11} & \textbf{85.86} & 11.6838 & 4.5887 \\
\bottomrule
\end{tabular}
}
\end{table}

\section{Discussion} \label{app:limitations}

\textbf{Limitations and Future Work.}
Our method is developed within prompt-based continual learning, where the downstream classifier plays an important but relatively underexplored role in continual OOD detection.
TA-KAN specifically addresses task-recency bias at the classifier level through adaptive RBF locality, rather than modifying the underlying feature representations.
While this design is well suited to scalable continual learning with pre-trained backbones, extending the proposed task-adaptive mechanism and geometry-guided scoring to other classifier architectures and non-prompt-based frameworks remains an interesting direction for future work.

\textbf{Broader Impact.}
This work advances out-of-distribution detection in continual learning (OOD-CL), enabling more reliable deployment of machine learning systems in open-world environments. By improving detection of unseen inputs without storing past data, the proposed rehearsal-free framework enhances safety and reduces overconfidence in applications such as healthcare, autonomous driving, and surveillance. Additionally, its privacy-preserving and scalable design makes it well-suited for real-world systems with evolving data streams.



\end{document}